\documentclass[lettersize,journal]{IEEEtran}
\usepackage{marvosym}
\usepackage{amsmath,amsfonts}
\usepackage{algorithm}
\usepackage{algorithmicx}
\usepackage{algpseudocode}
\usepackage{setspace}
\usepackage{array}
\usepackage{textcomp}
\usepackage{stfloats}
\usepackage{url}
\usepackage{verbatim}
\usepackage{graphicx}
\usepackage{color}
\usepackage{cite}
\usepackage{threeparttable}
\usepackage{booktabs}
\usepackage{multirow}
\usepackage{makecell}
\usepackage{minibox}
\usepackage{subcaption}
\usepackage[breaklinks,colorlinks, citecolor=blue]{hyperref}
\usepackage{amssymb}
\usepackage{tcolorbox}
\usepackage{colortbl}
\usepackage{pifont}
\usepackage{xcolor}
\usepackage{amsthm}
\definecolor{LightCyan}{rgb}{0.88,1,0.88}
\definecolor{LightRed}{rgb}{1,0.88,0.88}
\definecolor{grayish}{rgb}{0.95, 0.95, 0.95}
\definecolor{blackish}{rgb}{0.2, 0.2, 0.2}
\definecolor{beaublue}{rgb}{0.9, 0.95, 0.9}

\newtheorem{theorem}{Theorem}

\newtheorem{proposition}{Proposition}

\DeclareMathOperator*{\projj}{\text{$\Pi$}}

\newcommand{\methodname}{EvoUAD-AT}
\newcommand{\modname}{EvoAUM}
\newcommand{\R}{\mathbb{R}}
\newcommand{\E}{\mathbb{E}}
\newcommand{\KL}{\mathrm{KL}}
\newcommand{\vu}{\mathbf{u}}
\newcommand{\vm}{\mathbf{m}}
\newcommand{\vc}{\mathbf{c}}
\newcommand{\vp}{\mathbf{p}}
\newcommand{\vz}{\mathbf{z}}
\newcommand{\vs}{\mathbf{s}}
\newcommand{\vw}{\boldsymbol{\omega}}
\newcommand{\mB}{\mathbf{B}}
\newcommand{\mC}{\mathbf{C}}
\newcommand{\mI}{\mathbf{I}}
\newcommand{\hf}{\mathbf{\hat{f}}}
\newcommand{\cf}{\mathbf{\check{f}}}
\newcommand{\tF}{\widetilde F}
\newcommand{\algbox}[2]{\fcolorbox{white}{grayish}{\parbox[t]{\dimexpr\linewidth-\algorithmicindent*#1-2\fboxsep-2\fboxrule\relax}{#2}}}
\DeclareMathOperator{\diag}{diag}
\DeclareMathOperator{\clip}{clip}
\DeclareMathOperator{\orth}{orth}
\DeclareMathOperator{\tr}{tr}

\begin{document}

\title{Survival of the Fittest Perturbations: Uncertainty-Aware Evolution of Adversarial Distributions for Robust Generalization}

\author{Junhao~Dong,
    Zicheng Yu,
	Xinghua~Qu,
	Z. Jane Wang,~\IEEEmembership{Fellow,~IEEE},
	and~Yew-Soon~Ong\textsuperscript{\Letter},~\IEEEmembership{Fellow,~IEEE}
\thanks{Junhao Dong, Xinghua Qu, and Yew-Soon Ong are with the College of Computing and Data Science, Nanyang Technological University, Singapore, and also with the Center for Frontier AI Research, Agency for Science, Technology and Research, Singapore (e-mail: \{junhao003, xinghua001, asysong\}@ntu.edu.sg). Zicheng Yu is with the School of Information Management, Nanjing University, Nanjing, China (e-mail: yuzicheng@smail.nju.edu.cn). Z. Jane Wang is with the Department of Electrical and Computer Engineering, University of British Columbia, Vancouver, BC V6T 1Z4, Canada (e-mail: zjanew@ece.ubc.ca).}
\thanks{\Letter~denotes corresponding authors.}}

\markboth{IEEE Transactions on Evolutionary Computation}%
{Dong \MakeLowercase{\textit{et al.}}: Survival of the Fittest Perturbations: Uncertainty-Aware Evolution of Adversarial Distributions for Robust Generalization}


\maketitle

\begin{abstract}
Despite deep learning's broad successes across diverse scenarios, its vulnerability to adversarial examples impedes deployment, making adversarial training a leading approach to improve robustness. However, pointwise adversarial training maps each clean sample to a single adversary, overlooking surrounding adversarial variability and the associated shifts in feature statistics. To address these issues, we propose Evolutionary Uncertainty-Aware Distributional Adversarial Training (\methodname{}). Specifically, we design an evolutionary adversarial uncertainty modeling module that maintains an instance-wise search distribution over adversarial feature statistics around an adversarially generated anchor. Candidates are evaluated under the current model, selected by their threat to a benign reference, and used to adapt the distribution; strategy parameters are retained across epochs. To avoid anchoring on misclassified clean samples, we refine the alignment reference and match benign and adversarial distributions. Introspective gradient matching further regularizes input gradients without external supervision. Our analysis characterizes the elite fitness surrogate and gives a conditional certificate for latent adversarial features. Extensive evaluations across multiple benchmarks demonstrate that our approach consistently outperforms existing state-of-the-art methods in both clean accuracy and robustness, with thorough analyses confirming the effectiveness of each proposed component.

\end{abstract}

\begin{IEEEkeywords}
Adversarial training, evolution strategies, uncertainty modeling, distributional alignment, robust generalization.
\end{IEEEkeywords}

\vspace{-4mm}
\section{Introduction}
\IEEEPARstart{A}{lthough} Deep Neural Networks (DNNs) have revolutionized numerous fields with superior performance, a growing body of research has underscored a critical vulnerability of DNNs: their susceptibility to adversarial examples\textemdash inputs subtly modified with imperceptible perturbations \cite{SzegedyZSBEGF13}. These tailored examples can easily deceive human observers and mislead DNNs to output with high confidence, posing potential security risks. The growing vulnerabilities necessitate robust countermeasures against adversaries to safeguard the trustworthiness of DNNs in real-world applications.

\begin{figure}[t]
\centering
\begin{subcaptiongroup}
\phantomcaption\label{fig:1_1}
\phantomcaption\label{fig:1_2}
\phantomcaption\label{fig:1_3}
\end{subcaptiongroup}
\includegraphics[width=\linewidth]{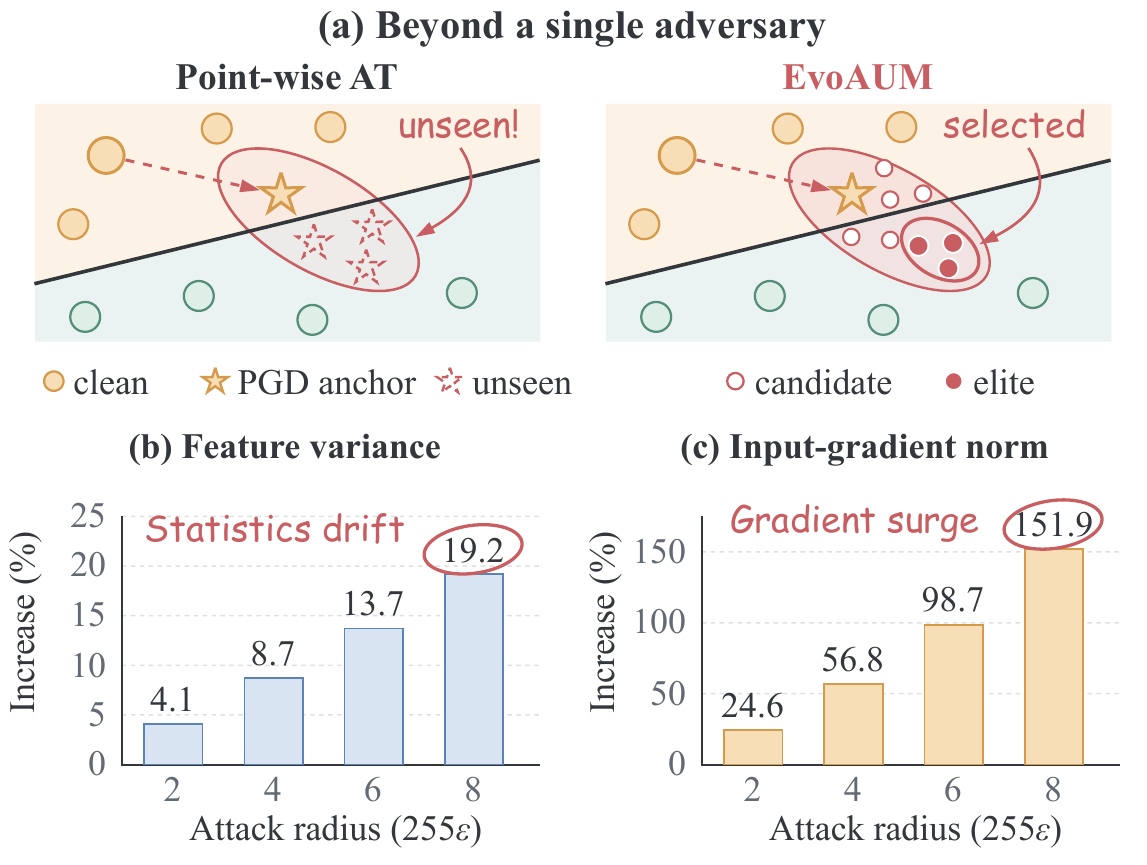}
\vspace{-7mm}
\caption{Motivation for evolutionary distributional training. (a) Point-wise AT trains on one PGD anchor; \modname{} explores its feature-statistic neighborhood and selects threatening candidates (schematic). Under stronger PGD attacks, the standard PGD-AT model~\cite{MadryMSTV18} exhibits larger increases in (b) feature variance and (c) input-gradient norm relative to clean inputs.}
\label{fig:1}
\vspace{-3mm}
\end{figure}

Among various defense methods against adversarial examples, adversarial training has emerged as the most effective one for improving intrinsic network robustness via augmenting adversaries into training samples \cite{GoodfellowSS15, MadryMSTV18, zhang2019theoretically, RadeM22, dong2023enemy}. These works primarily rely on a point-by-point augmentation strategy, wherein each clean example is transformed into a single adversarial counterpart during the training stage. However, this deterministic training paradigm overlooks the multitude of latent adversarial examples around the decision boundaries, which compromises the performance of classification models \cite{BuiLT0P22}. Such an oversight can lead to an overfitted classifier that suffers from a distribution shift away from latent adversaries, resulting in suboptimal decision boundaries. Furthermore, adversaries can induce a significant disruption to the feature-level statistical information within the target adversary domain, exacerbating the complexity of modeling adversarial distributions.

Hence, we hypothesize that classifiers trained via a point-by-point adversarial strategy suffer from overfitting to isolated adversaries, leading to poor generalization across surrounding latent perturbations (Figure~\ref{fig:1_1}). Moreover, the underlying distribution mismatch between clean and adversarial domains further exacerbates feature-level statistical deviations. To verify this, we analyze the feature variance increment from clean examples to their adversarial counterparts on a PGD-AT model~\cite{MadryMSTV18} (Figure~\ref{fig:1_2}). The positive correlation between feature variance escalation and perturbation radius confirms that stronger adversaries induce pronounced statistical shifts in feature representations. Similarly, the input-gradient norm with respect to the cross-entropy loss rises markedly as the perturbation budget enlarges (Figure~\ref{fig:1_3}), demonstrating heightened local instability and vulnerability to potent attacks.

To mitigate the potential overfitting to a single adversary generated through the point-by-point strategy from each clean example, we propose \textit{Evolutionary Adversarial Uncertainty Modeling} (\textbf{\modname{}}). Around a PGD-generated feature anchor, \modname{} samples candidate feature statistics, evaluates their threat under the current model, and updates an instance-wise search distribution through rank-based selection. Intermediate PGD iterates provide the prior scale and search directions, while historical information is retained as strategy parameters and re-evaluated through newly sampled candidates at each visit. Recognizing the detrimental impact brought by prediction alignment between adversaries and misclassified clean examples in adversarial training, we design a distributional refinement scheme to correct the alignment reference (\textit{i.e.}, clean examples) based on the statistical proximity to clean examples during adversarial training. Consequently, adversarial training is reframed into a distribution-to-distribution matching framework between clean and adversarial domains. This framework, termed \textit{Evolutionary Uncertainty-Aware Distributional Adversarial Training} (\textbf{\methodname{}}), transcends traditional instance-specific schemes by considering both the underlying adversaries and their benignly refined counterparts to capture more generalizable and robust behavior. To enforce prediction invariance against adversarial perturbations, we develop an introspective gradient matching approach by aligning input gradients across clean and adversarial domains without incorporating external models.

We conduct extensive experiments across four standard benchmarks, demonstrating that \methodname{} consistently outperforms state-of-the-art adversarial training baselines across diverse network backbones and perturbation budgets. Moreover, our method scales effectively with auxiliary synthetic data and large-capacity architectures, and integrates seamlessly into single-step adversarial training to substantially mitigate catastrophic overfitting with only modest computational overhead. Comprehensive ablation studies and sensitivity analyses further isolate the individual contributions of our evolutionary feature search and prior-guided initialization.

Our contributions can be summarized as follows:

\begin{itemize}
\item By investigating the limited generalizability towards underlying adversarial examples stemming from point-by-point adversary generation, we propose \modname{}, which combines feature-statistic uncertainty with current-model fitness evaluation, rank-based selection, distribution adaptation, and cross-epoch strategy memory for distributional adversarial training.

\item To mitigate the detrimental impact of prediction alignment on misclassified examples, we distributionally refine the alignment reference within a distribution-to-distribution matching scheme and introduce an introspective gradient matching mechanism to enforce prediction invariance without external supervision. Moreover, for a fixed benign reference, we characterize the elite fitness surrogate and derive a conditional robustness bound for latent adversarial features.

\item We provide a theoretical analysis showing that modeling instance-wise feature uncertainty via rank-based evolutionary search provably expands the local adversarial coverage around point-wise PGD anchors, thereby tightening the generalization bound against unforeseen latent perturbations without sacrificing clean accuracy.
\end{itemize}


\section{Related Works}
\subsection{Adversarial Attack}
Adversarial attacks are deliberately crafted perturbations applied to the model input, typically of small magnitude, which induce mispredictions at inference \cite{SzegedyZSBEGF13, GoodfellowSS15, su2024multi, li2025map}. Szegedy \textit{et al.} \cite{SzegedyZSBEGF13} first exposed this vulnerability and coined ``adversarial examples,'' while Madry \textit{et al.} \cite{MadryMSTV18} used projected gradient descent (PGD) to search for high-loss inputs within a prescribed perturbation budget. Beyond focusing solely on the final adversarial example, some works analyze the intermediate iterates produced along the attack trajectory for diagnostic purposes. In this paper, these iterates estimate the instance-level prior scale and guide evolutionary sampling in feature-statistic space. In parallel, a substantial body of work studies black-box transferability \cite{zhu2022toward}. In our white-box evaluations, the attacker has access to the model parameters and can compute gradients with respect to the input.

\subsection{Adversarial Training}
Given the security risks of adversarial examples, adversarial training \cite{MadryMSTV18, zhang2019theoretically, 0001ZY0MG20, RadeM22, huang2023fast} remains the most effective robustness approach, augmenting training with adversarial examples. To benchmark the clean–robustness trade-off, Zhang \textit{et al.} \cite{zhang2019theoretically} formulated adversarial training as prediction alignment between each clean sample and its adversarial counterpart. Rade \textit{et al.} \cite{RadeM22} observed unnecessary margin growth along adversarial directions and improved the trade-off by introducing examples with adversarial labels. To correct misalignment when clean samples are misclassified, Dong \textit{et al.} \cite{dong2023enemy} proposed inverse adversaries that reset the alignment reference to the clean example. However, fixing a single (locally invariant) reference harms the accuracy–robustness balance \cite{pang2022robustness}. We instead propose a distributional refinement that dynamically rectifies the prediction-alignment reference via statistical proximity to clean data, leveraging distributional modeling to mitigate the inductive bias toward local invariance.

\subsection{Deep Uncertainty Learning}
Uncertainty Learning has advanced a wide range of deep learning applications. Variational Autoencoders (VAEs) model data distributions to yield robust representations. Shi \textit{et al.} \cite{shi2019probabilistic} convert deterministic face embeddings into probabilistic distributions to filter low-quality inputs. Li \textit{et al.} \cite{LiDGLSD22} characterize statistical shifts in out-of-distribution data and approximate distributions for each feature statistic. However, their uncertainty estimates are largely batch-level, discarding instance-specific attributes. \modname{} uses current-model trajectories to guide sampling and retains only strategy parameters across epochs. Note that every adversarial candidate is newly generated and evaluated under the current model.

Incorporating uncertainty into adversarial training has also been explored \cite{dong2020adversarial, 10354457, 10177878}. Dong \textit{et al.} \cite{10177878} demonstrated the prevalence of latent adversaries and proposed distributional adversarial training that augments adversaries with estimated feature-level statistics. Zhao \textit{et al.} \cite{10354457} linked adversary complexity to robustness and introduced a variational defense using Bayesian inference to approximate adversarial distributions. Most distributional adversarial training aligns an adversarial distribution to a fixed clean example at the prediction level (distribution-to-point), a choice prone to misclassification and an inductive bias toward local invariance, yielding suboptimal decision boundaries. We instead adopt distributional prediction matching, dynamically refining the alignment reference via integrating uncertainty over feature statistics for more generalizable adversarial training.

Evolution strategies adapt a search distribution through population evaluation, selection, and recombination~\cite{hansen2016tutorial,ros2008sep}. Guided ES uses surrogate directions to shape sampling~\cite{maheswaranathan2019guided}. We combine these principles with the feature-statistic representation of AUM: PGD supplies an anchor and directional information, and \modname{} evolves a distribution of candidate features around that anchor. Throughout this paper, AUM denotes the random augmentation module, \modname{} the evolutionary module, and \methodname{} the complete framework.

\section{Proposed Method}
\begin{figure*}[!t]
\centering
\vspace{-4mm}
\includegraphics[width=0.95\linewidth]{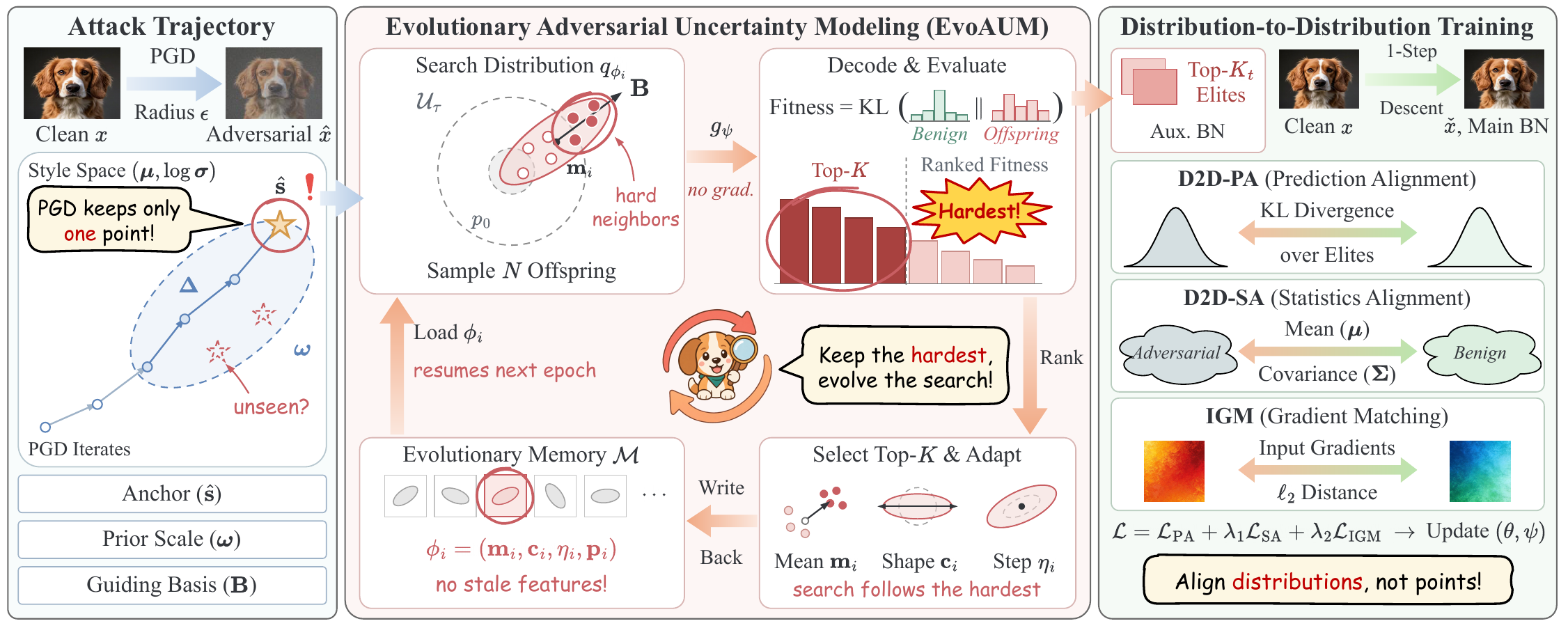}
\vspace{-1mm}
\caption{Pipeline and components of our EvoUAD-AT. Evolutionary adversarial uncertainty modeling augments feature statistics, distributional alignment matches predictions, feature statistics across branches, and gradient alignment matches input gradients.}
\label{fig:method}
\end{figure*}

In this section, we first formalize adversarial training. We then present \textit{Evolutionary Uncertainty-Aware Distributional Adversarial Training} (\textbf{\methodname{}}), which replaces the pointwise clean-to-adversary alignment of standard adversarial training with the distribution-to-distribution matching between the clean and adversarial domains. The framework comprises \modname{} for evolutionary feature search, Distributional Alignment Refinement (DAR) for the benign reference and domain alignment, and Introspective Gradient Matching (IGM) for input-gradient regularization. PGD still generates the adversarial anchor; the evolutionary operations belong to \modname{}. Figure~\ref{fig:method} illustrates the feature augmentation and alignment operations in EvoUAD-AT. Algorithm~\ref{alg:1} gives the training procedure for \methodname{}.

\subsection{Problem Formulation}
The main goal of adversarial training is to build robustness for a certain model while maintaining its natural performance, which can be achieved by augmenting adversaries during training. For classification, we consider a DNN-based $C$-category classifier that integrates a feature extractor, denoted as $f_{\boldsymbol{\theta}}: \mathcal{X} \rightarrow \mathcal{F}$, and its corresponding classification head, represented by $g_{\boldsymbol{\psi}}: \mathcal{F} \rightarrow \mathbb{R}^{C}$ with network parameters $\boldsymbol{\theta}$ and $\boldsymbol{\psi}$, respectively. $\mathcal{F}$ represents the intermediate feature space $\mathbb{R}^{D \times H \times W}$, which is characterized by $D$ channels of $H \times W$ feature maps. Given a specific dataset $\left( \mathbf{x}, y\right)\sim \mathcal{D}$, a classic adversarial training baseline (TRADES \cite{zhang2019theoretically}) optimizes a surrogate upper bound of the robust risk under the $\ell_{\infty}$-norm threat model, which can be conducted by optimizing the following minimax problem:
\begin{equation}
\begin{aligned}
    \min\limits_{\boldsymbol{\theta}, \boldsymbol{\psi}}& \mathbb{E}_{\left( \mathbf{x}, y\right)\sim \mathcal{D} }  \Big[ \mathcal{L}_{\text{CE}} \left( g_{\boldsymbol{\psi}}(f_{\boldsymbol{\theta}} \left( \mathbf{x}\right)) , y \right) + \\
    &\beta \cdot  \max\limits_{\left\| \boldsymbol{\delta} \right\|_{\infty}<\epsilon}\mathcal{L}_{\text{KL}}\left( g_{\boldsymbol{\psi}}(f_{\boldsymbol{\theta}} \left( \mathbf{x}\right))  \| g_{\boldsymbol{\psi}}(f_{\boldsymbol{\theta}}\left( \mathbf{x} + \boldsymbol{\delta} \right)) \right)   \Big],
    \label{eq:1}
\end{aligned}
\end{equation}
where $\mathcal{L}_{\text{CE}}$ and $\mathcal{L}_{\text{KL}}$ represent the Cross-Entropy (CE) loss and the Kullback--Leibler (KL) divergence between the softmax predictions of the supplied logits. $\beta$ controls the trade-off between natural performance and adversarial robustness. Adversarial perturbation $\boldsymbol{\delta}$ is bounded within the $\ell_{\infty}$-norm radius $\epsilon$. The outer minimization optimizes the natural empirical risk along with the prediction alignment between clean and adversarial examples over network parameters $\boldsymbol{\theta}$. Meanwhile, the inner optimization aims to find the worst-case adversarial examples $\mathbf{\hat{x}}=\mathbf{x}+\boldsymbol{\delta}$, which deviates the target classification model to make wrong predictions. The inner optimization uses Projected Gradient Descent (PGD)~\cite{MadryMSTV18}. Writing its attack loss as $\ell_{\mathrm{atk}}$, the update is:
\begin{equation}
\hat{\mathbf{x}}^{(a+1)}=\Pi_{\mathbb{B}(\mathbf{x},\epsilon)}\left(\hat{\mathbf{x}}^{(a)}+\alpha\operatorname{sgn}\left(\nabla_{\hat{\mathbf{x}}^{(a)}}\ell_{\mathrm{atk}}(\hat{\mathbf{x}}^{(a)})\right)\right).
\label{eq:2}
\end{equation}
For the TRADES baseline in Eq.~\eqref{eq:1}, $\ell_{\mathrm{atk}}$ is the prediction KL divergence. The attack loss for our training runs is specified in the implementation details; it is separate from the feature-level fitness used by \modname{} below.

where $\operatorname{sgn}(\cdot)$ represents the sign function, and $\alpha$ denotes the step size of iterative gradient ascent. $\Pi_{\mathbb{B}(\mathbf{x}, \epsilon)}(\cdot)$ is the $\ell_{\infty}$-norm projection function with radius $\epsilon$ around clean examples $\mathbf{x}$ for the visual imperceptibility of adversarial perturbation. The random start strategy is utilized for initialization $\mathbf{\hat{x}}^{(0)} = \mathbf{x} + 0.001\cdot\mathcal{N}(0,1)$ to escape the local maxima. For brevity, we denote the clean feature as $\mathbf{f}=f_{\boldsymbol{\theta}}(\mathbf{x})$ and its corresponding adversarial feature as $\mathbf{\hat{f}}=f_{\boldsymbol{\theta}}(\mathbf{\hat{x}})$. Note that adversary generation is an $n$-step iterative process involving a series of intermediate results $\mathcal{I} = \{ \mathbf{\hat{x}}^{(i)} \}_{i=1}^{n-1}$ that are also threatening to the target model. Due to the non-linear dynamics of decision boundaries, DNNs are also potentially vulnerable to these intermediate adversaries, which necessitates attention during robustness establishment. In short, we primarily consider adversarial training for classification models under the $\ell_{\infty}$-norm setting, if not specified otherwise.

\subsection{Evolutionary Adversarial Uncertainty Modeling}
\label{sec:AUM}
As we previously mentioned, prior adversarial training methods mainly relied on adversaries generated via a point-to-point scheme, yet they inherently overlook the stochastic nature of adversary generation. This oversight can lead to suboptimal decision boundaries for a wide spectrum of underlying adversaries. Moreover, we have shown the disruptive potential of adversaries towards their statistical properties at the feature level (see Figure \ref{fig:1}), complicating the robust modeling of adversarial distributions. To address these issues, \modname{} maintains and adapts a search distribution over feature statistics, using current-model fitness to select threatening candidates for training.

\begin{figure}[t]
\centering
\begin{subfigure}[t]{0.32\linewidth} 
    \centering
    \includegraphics[width=1\linewidth]{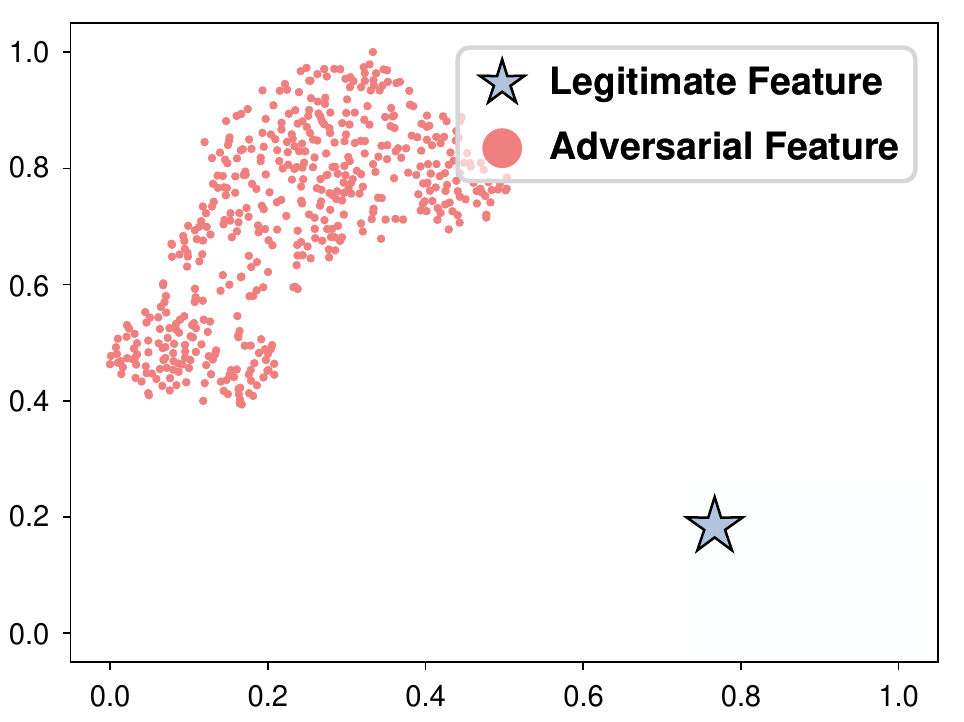}
    \vspace{-6mm}
    \caption{}
    \label{fig:2_1}
\end{subfigure} 
\begin{subfigure}[t]{0.32\linewidth}
    \centering
    \includegraphics[width=1\linewidth]{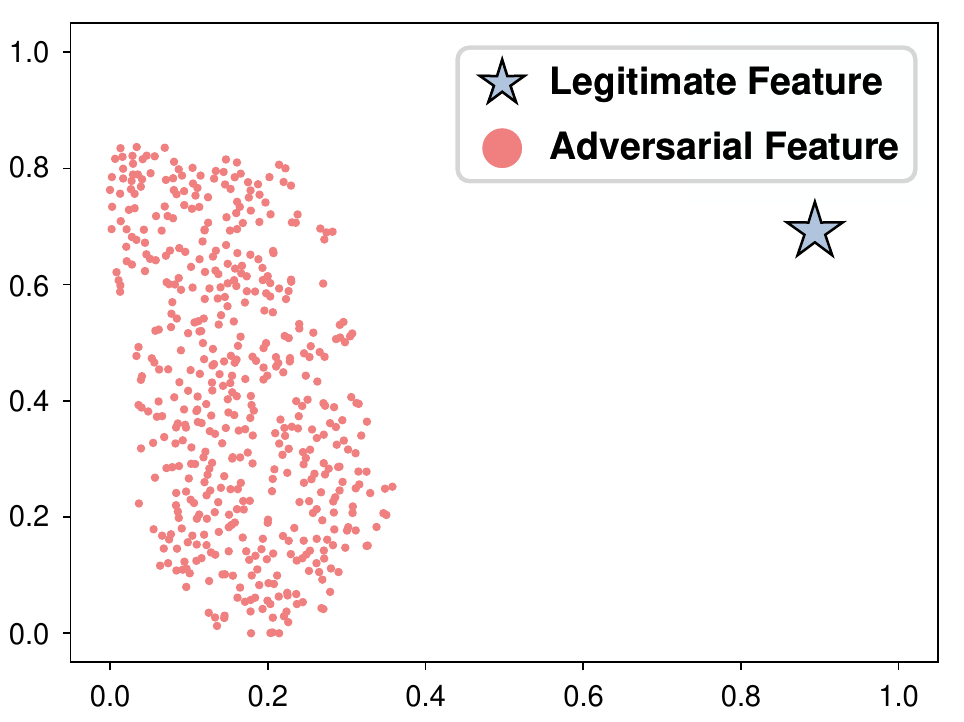}
    \vspace{-6mm}
    \caption{}
    \label{fig:2_2}
\end{subfigure}
\begin{subfigure}[t]{0.32\linewidth}
    \centering
    \includegraphics[width=1\linewidth]{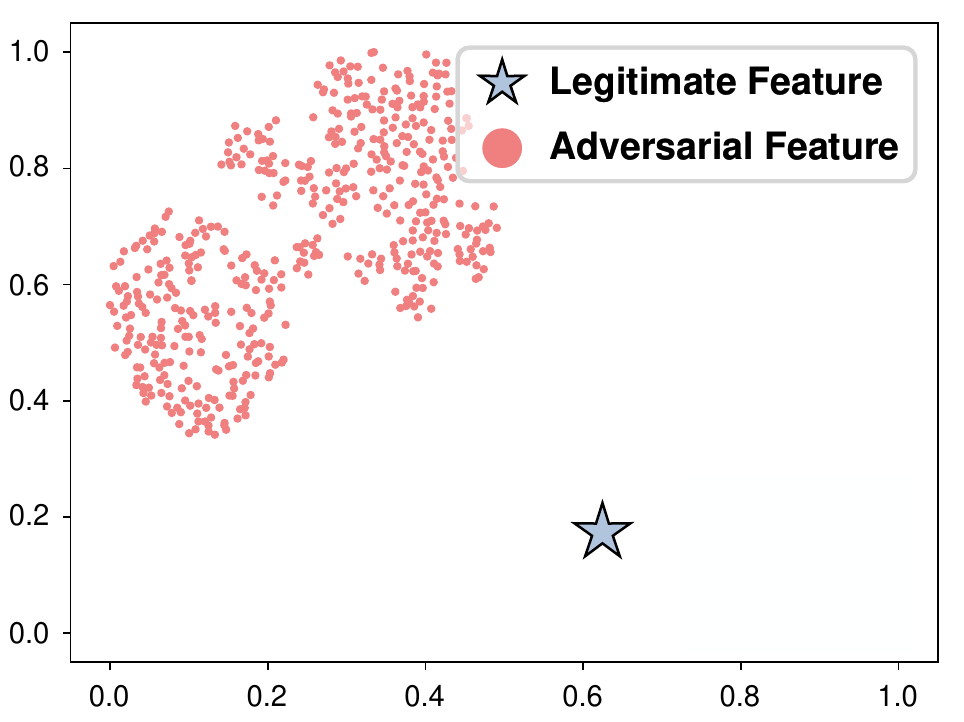}
    \vspace{-6mm}
    \caption{}
    \label{fig:2_3}
\end{subfigure}
\begin{subfigure}[t]{0.32\linewidth} 
    \centering
    \includegraphics[width=1\linewidth]{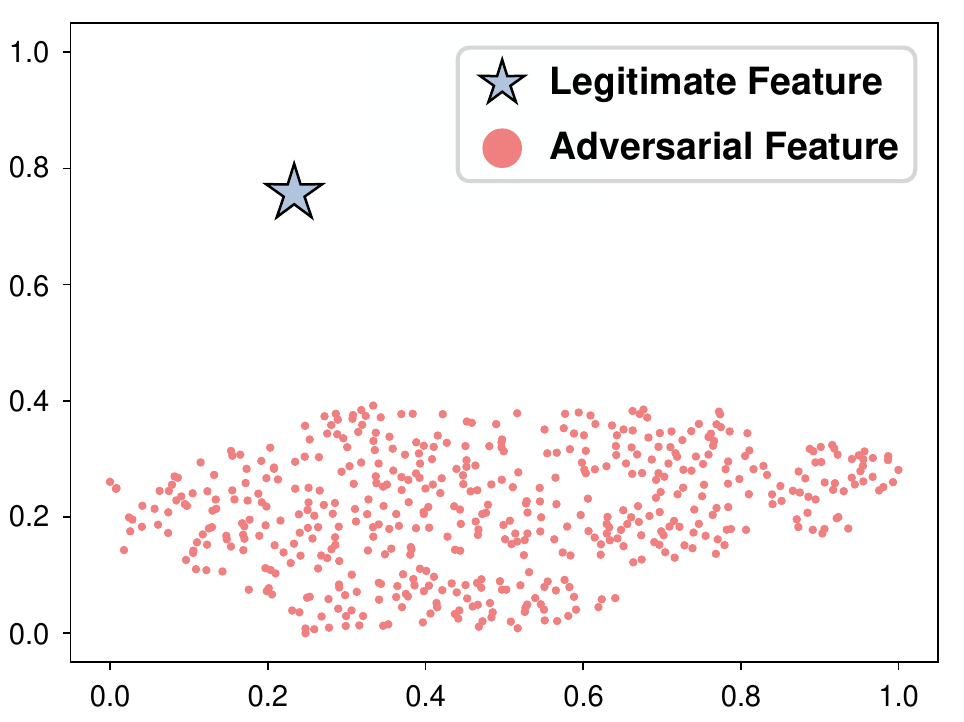}
    \vspace{-6mm}
    \caption{}
    \label{fig:2_4}
\end{subfigure} 
\begin{subfigure}[t]{0.32\linewidth}
    \centering
    \includegraphics[width=1\linewidth]{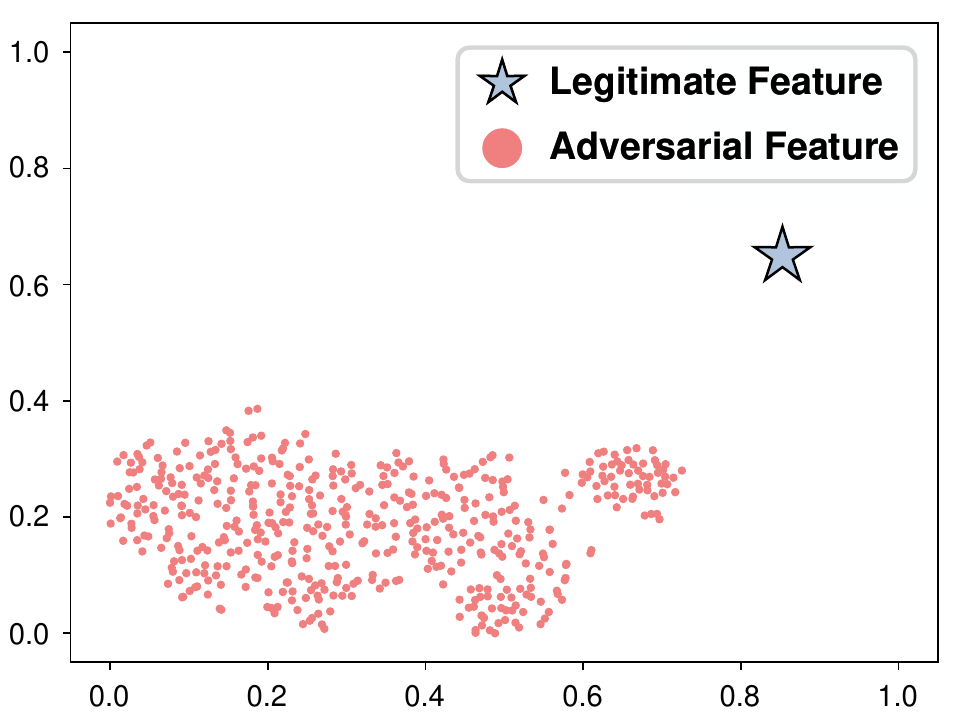}
    \vspace{-6mm}
    \caption{}
    \label{fig:2_5}
\end{subfigure}
\begin{subfigure}[t]{0.32\linewidth}
    \centering
    \includegraphics[width=1\linewidth]{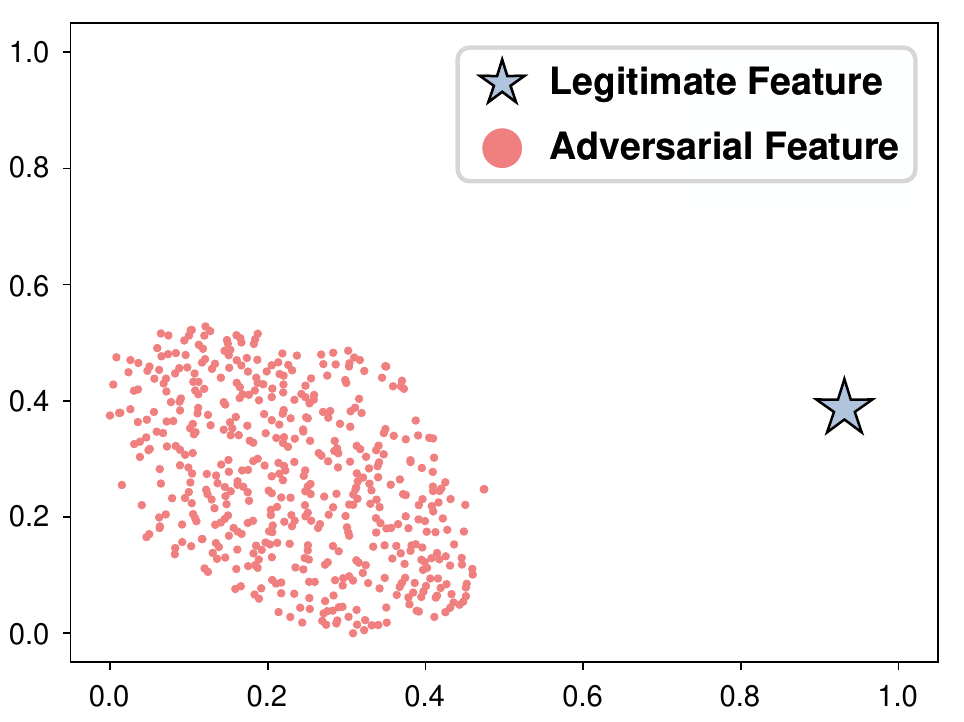}
    \vspace{-6mm}
    \caption{}
    \label{fig:2_6}
\end{subfigure}
\vspace{-2mm}
\caption{t-SNE visualization of legitimate feature and its 500 adversarial counterparts generated via the random start strategy on CIFAR-10 \cite{krizhevsky2009learning}. Each figure refers to a specific clean input.
}
\label{fig:2}
\vspace{-5mm}
\end{figure}

\begin{figure*}[!t]
\centering
\begin{subfigure}[t]{0.3\linewidth} 
    \centering
    \includegraphics[width=1\linewidth]{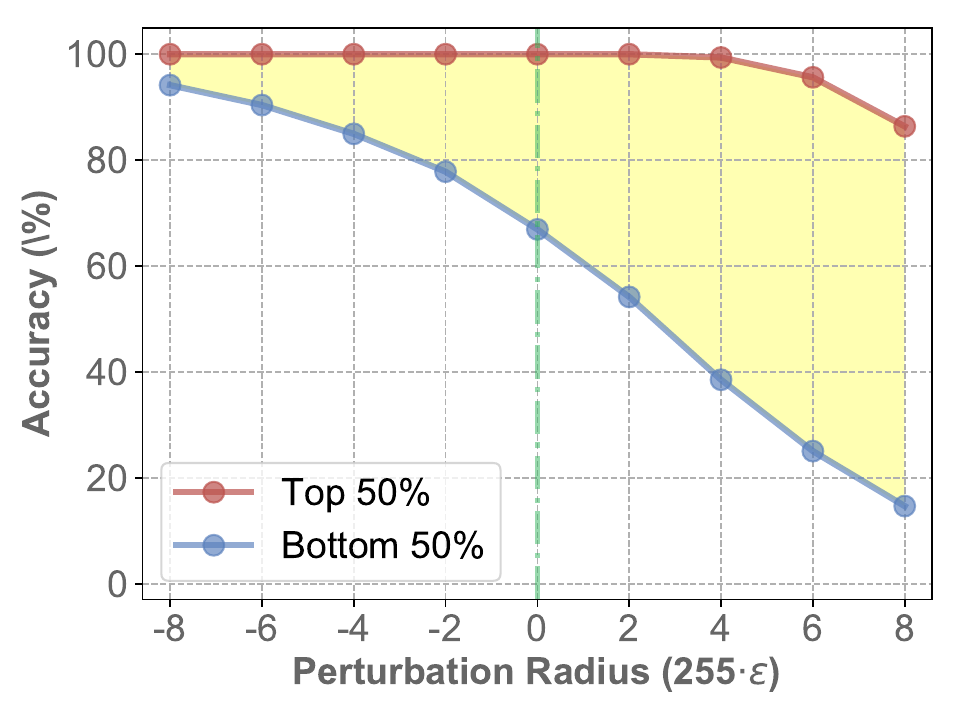}
    \vspace{-7mm}
    \caption{Robust Accuracy}
    \label{fig:3_1}
\end{subfigure} 
\hfill
\begin{subfigure}[t]{0.3\linewidth}
    \centering
    \includegraphics[width=1\linewidth]{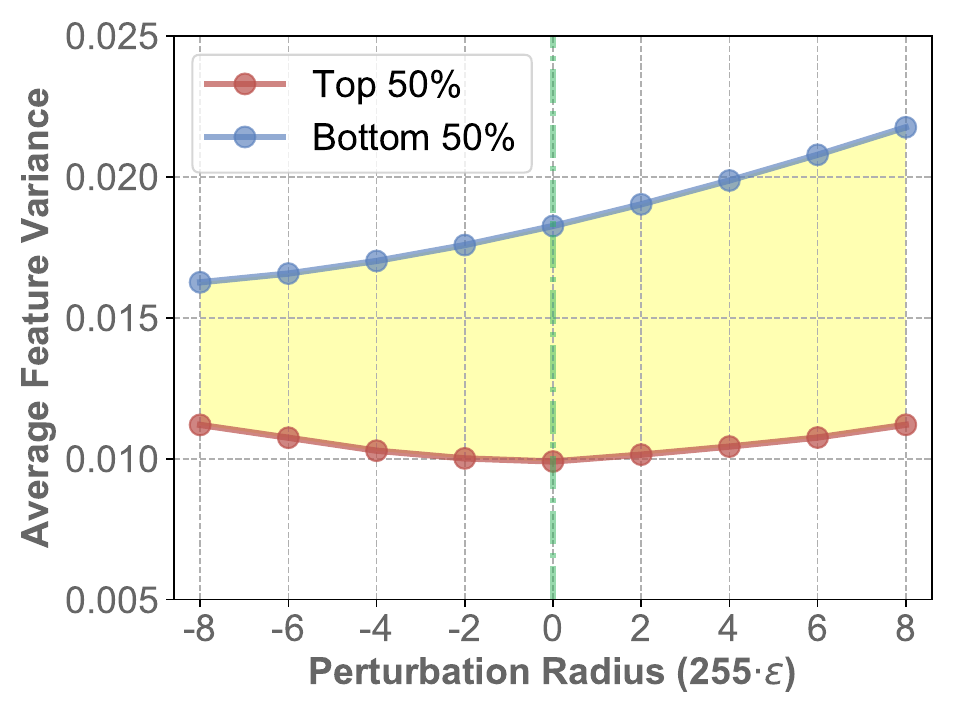}
    \vspace{-7mm}
    \caption{Feature Variance}
    \label{fig:3_2}
\end{subfigure}
\hfill
\begin{subfigure}[t]{0.3\linewidth}
    \centering
    \includegraphics[width=1\linewidth]{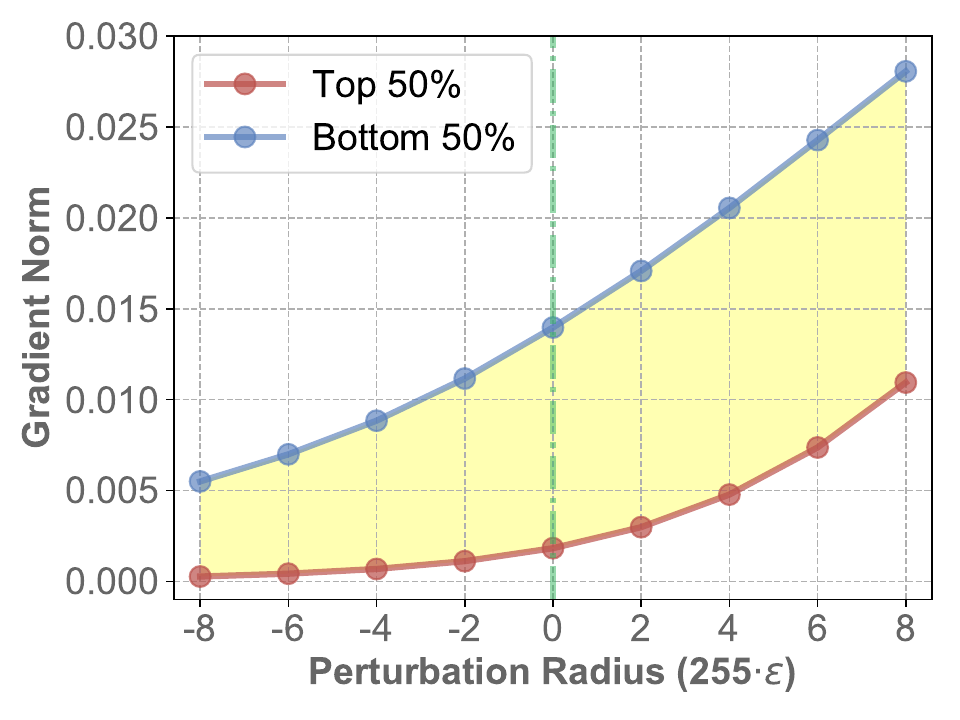}
    \vspace{-7mm}
    \caption{Gradient Norm}
    \label{fig:3_3}
\end{subfigure}
\vspace{-2mm}
\caption{(a) Average robust accuracy, (b) feature variance, and (c) gradient norm under different attack strengths for an adversarially trained model (PGD-AT \cite{MadryMSTV18}). Test samples are sorted by cross-entropy (ascending) and split into two equal halves. Negative perturbation radii denote benign refinement by flipping the attack gradient sign during adversary generation. The \textcolor[RGB]{79,180,121}{green} line marks clean samples.
}
\label{fig:3}
\vspace{-3mm}
\end{figure*}

We parameterize candidate perturbations by channel-wise feature statistics. Figure~\ref{fig:2} illustrates the variation among adversarial features for individual inputs. \modname{} samples offsets from a Gaussian proposal in feature-statistic space. For a PGD anchor feature $\hf\in\R^{D\times H\times W}$, the channel means and inter-channel covariances are
\begin{equation}
	\begin{aligned}
		\boldsymbol{\hat{\mu}}_i = \frac{1}{HW} \sum_{h=1}^{H} \sum_{w=1}^{W} \mathbf{\hat{f}}_{i, h, w},
		\label{eq:stats_1}
	\end{aligned}
\end{equation}
\vspace{-2mm}
\begin{equation}
	\begin{aligned}
		\boldsymbol{\hat{\Sigma}}_{i,j} = \frac{1}{HW} \sum_{h=1}^{H} \sum_{w=1}^{W} \left(\mathbf{\hat{f}}_{i, h, w} - \boldsymbol{\hat{\mu}}_i \right) \left( \mathbf{\hat{f}}_{j, h, w} - \boldsymbol{\hat{\mu}}_j \right),
		\label{eq:stats_2}
	\end{aligned}
\end{equation}

where $i$ and $j$ index channels. The same definitions apply to any feature tensor $f$. For normalization and logarithms, let $\sigma_j(f)=(\Sigma_{j,j}(f)+\iota)^{1/2}$ with $\iota>0$. Mean and covariance summaries are also used by DAR in Section~\ref{sec:MethodDAR}.

\noindent \textbf{Representation and current-model prior.} Following the feature-statistic view of adaptive instance normalization~\cite{huang2017arbitrary}, define the style vector and re-styling map
\begin{align}
\vs(f)&=[\boldsymbol{\mu}(f);\log\boldsymbol{\sigma}(f)]\in\R^d,\qquad d=2D,\label{eq:evo-style}\\
\mathcal{T}(f;\vs')&=\exp(\vs'_\sigma)\odot\frac{f-\boldsymbol{\mu}(f)}{\boldsymbol{\sigma}(f)}+\vs'_\mu.\label{eq:evo-restyle}
\end{align}
Here $\vs'_\mu$ and $\vs'_\sigma$ are the mean and log-standard-deviation segments. Let $\hat\vs=\vs(\hf)$. For a multi-step attack, using the last $\kappa$ iterates of the current PGD trajectory, write $\vs^{(a)}=\vs(f_{\boldsymbol{\theta}}(\hat{\mathbf{x}}^{(a)}))$ and $\bar\vs=\kappa^{-1}\sum_{a=n-\kappa+1}^n\vs^{(a)}$. We estimate a positive prior scale as
\begin{equation}
\vw=\max\left\{\left[\frac1\kappa\sum_{a=n-\kappa+1}^{n}(\vs^{(a)}-\bar\vs)^{\odot2}\right]^{\odot\frac12},\omega_{\min}\right\}.
\label{eq:5}
\end{equation}
All operations are elementwise and $\omega_{\min}>0$. An individual is a whitened offset $\vu\in\R^d$, decoded into an adversarial feature by
\begin{equation}
\hf(\vu)=\mathcal{T}(\hf;\hat\vs+\vw\odot\vu).
\label{eq:7}
\end{equation}
Both AUM and \modname{} augment features. \modname{} selects candidates by fitness and adapts the offset distribution.

\noindent \textbf{Fitness and feasible candidates.} Let $\cf=f_{\boldsymbol{\theta}}(\mathbf{\check{x}})$ be the benign reference obtained by Eq.~\eqref{eq:8}, and let $\pi_{\boldsymbol{\psi}}(f)=\operatorname{softmax}(g_{\boldsymbol{\psi}}(f))$. Candidate threat is measured by
\begin{equation}
F_i(\vu)=\KL\big(\pi_{\boldsymbol{\psi}}(\cf_i)\,\|\,\pi_{\boldsymbol{\psi}}(\hf_i(\vu))\big).
\label{eq:evo-fitness}
\end{equation}
Larger values indicate greater prediction disagreement with the reference. Candidates are restricted to $\mathcal{U}_\tau=\{\vu:\|\vu\|_2\le\tau\sqrt d\}$ by radial repair
\begin{equation}
\Pi_\tau(\vu)=\vu\min\{1,\tau\sqrt d/\|\vu\|_2\},\qquad \Pi_\tau(\mathbf{0})=\mathbf{0}.
\label{eq:evo-repair}
\end{equation}
The input-space PGD budget $\epsilon$ and the feature-space search radius $\tau$ have distinct roles; a decoded feature need not correspond to an image within the input threat set.

\noindent \textbf{Trajectory-guided variation.} For each training instance $i$, we maintain $\phi_i=(\vm_i,\vc_i,\eta_i,\vp_i)$, where $\vm_i$ is the search mean, $\vc_i>0$ a diagonal variance factor, $\eta_i>0$ a step size, and $\vp_i$ an evolution path. The Gaussian proposal is
\begin{equation}
\scalebox{0.9}{$\displaystyle
q_{\phi_i}=\mathcal{N}(\vm_i,\mC_i),\quad
\mC_i=\eta_i^2\left(
\gamma\diag(\vc_i)+
\frac{1-\gamma}{r_i}\mB_i\mB_i^\top
\right)
$}
\label{eq:evo-cov}
\end{equation}
The orthonormal columns of $\mB_i$ span whitened displacements from the current attack trajectory:
\begin{equation}
\begin{aligned}
\boldsymbol{\Delta}_i^{(a)}&=(\vs_i^{(a)}-\vs_i^{(a-1)})\oslash\vw_i,\\
\mB_i&=\orth\big([\boldsymbol{\Delta}_i^{(n-\kappa+2)},\ldots,\boldsymbol{\Delta}_i^{(n)}]\big).
\end{aligned}
\label{eq:6}
\end{equation}
A thin QR decomposition yields rank $r_i\le\kappa-1$. This guided-ES construction~\cite{maheswaranathan2019guided} uses the trajectory as directional information. Intermediate iterates also determine $\vw_i$, but are not pooled with stale adversarial features. We sample $N$ offspring by
\begin{equation}
\scalebox{0.9}{$\displaystyle
\vu_{i,k}=\Pi_\tau\left(
\vm_i+\eta_i\left[
\sqrt\gamma\,\vc_i^{\odot\frac12}\odot\vz_{i,k}
+\sqrt{\frac{1-\gamma}{r_i}}\mB_i\boldsymbol{\zeta}_{i,k}
\right]\right)
$}
\label{eq:evo-sample}
\end{equation}
where $\vz_{i,k}\sim\mathcal{N}(\mathbf{0},\mI_d)$ and $\boldsymbol{\zeta}_{i,k}\sim\mathcal{N}(\mathbf{0},\mI_{r_i})$ are independent. If $r_i=0$, we set $\gamma=1$ and omit the low-rank terms in Eqs.~\eqref{eq:evo-cov} and~\eqref{eq:evo-sample}.

\noindent \textbf{Selection and adaptation.} The candidates are evaluated under the current model without gradient tracking or BN running-statistics updates. Rank them so that $F_i(\vu_{i,1:N})\ge\cdots\ge F_i(\vu_{i,N:N})$. For the top $K$ parents, use normalized log-rank weights $w_k\propto\ln(K+1/2)-\ln k$, with $\sum_{k=1}^K w_k=1$ and $K_{\mathrm{eff}}=(\sum_k w_k^2)^{-1}$. Following separable covariance adaptation and cumulative step-size adaptation~\cite{ros2008sep,hansen2016tutorial}, the updates are
\begin{align}
\vm_i&\leftarrow\Pi_\tau\left(\vm_i+c_m\sum_{k=1}^{K}w_k(\vu_{i,k:N}-\vm_i)\right),\label{eq:evo-mean}\\
\vc_i &\leftarrow
\scalebox{0.9}{$\displaystyle
\clip_{[c_{\min},c_{\max}]}\left(
(1-c_d)\vc_i+
c_d\vc_i\odot\sum_{k=1}^{K}
w_k\vz_{i,k:N}^{\odot2}
\right)
$}
\label{eq:evo-diag}\\
\vp_i&\leftarrow(1-c_s)\vp_i+\sqrt{c_s(2-c_s)K_{\mathrm{eff}}}\sum_{k=1}^Kw_k\vz_{i,k:N},\label{eq:evo-path}\\
\eta_i&\leftarrow\clip_{[\eta_{\min},\eta_{\max}]}\left(\eta_i\exp\left[\frac{c_s}{d_s}\left(\frac{\|\vp_i\|_2}{\chi_d}-1\right)\right]\right),\label{eq:evo-step}
\end{align}
where $\chi_d=\E\|\mathcal{N}(\mathbf{0},\mI_d)\|_2$ and $\vz_{i,k:N}$ is the original isotropic noise associated with the selected offspring. The variance and path updates use this noise component. Positive lower bounds on $\eta_i$, $\vc_i$, and $\gamma$ keep the Gaussian proposal covariance nonsingular. The $K_t\le K$ highest-ranked candidates supply the network-training loss in Eq.~\eqref{eq:9}.

\noindent \textbf{Evolutionary memory and cost.} An instance-indexed table stores only $\phi_i$, initialized to $(\mathbf{0},\mathbf{1},1,\mathbf{0})$. At each visit we recompute $\hat\vs_i$, $\vw_i$, and $\mB_i$ using the current model, load $\phi_i$, perform $G$ generations, and write back the updated strategy. This transfers search information in whitened coordinates across epochs without reusing old adversarial samples or their fitness. It requires $3d+1$ scalars per instance (approximately 77 MB in FP16 for 50,000 CIFAR-10 instances at $d=256$). Candidate evaluation uses batched forwards through $g_{\boldsymbol{\psi}}$; only selected features are decoded again with gradients for the network update.

\noindent \textbf{Reference configuration.} The method configuration uses $N=8$, $K=4$, $K_t=2$, $G=1$, $\kappa=5$, $\gamma=0.5$, and $\tau=1.5$, with $[\eta_{\min},\eta_{\max}]=[0.3,3]$, $[c_{\min},c_{\max}]=[0.1,10]$, $c_m=1$, and $c_d=0.1$; $c_s$ and $d_s$ follow~\cite{hansen2016tutorial}. The equations describe elite-only training.

With initial strategy $(\mathbf{0},\mathbf{1},1,\mathbf{0})$, $\gamma=1$, uniform weights over all $N$ candidates, no adaptation, and $\tau=\infty$, Eq.~\eqref{eq:7} reduces to random Gaussian offsets in style space. This shares AUM's random-augmentation mechanism, with additive mean shifts and first-order Gaussian standard-deviation shifts. \modname{} perturbs standard deviations in log space and estimates the search scale from the current PGD trajectory.

\subsection{Distributional Alignment Refinement}
\label{sec:MethodDAR}

With the evolutionary candidates selected, DAR supplies the benign reference and aligns the two domains for robust, generalizable decision boundaries. Prior work typically uses \emph{prediction alignment} to match each adversary to its clean counterpart; however, aligning multiple adversaries to a fixed reference induces local invariance, worsening the accuracy–robustness trade-off \cite{pang2022robustness}. Moreover, when the clean reference is misclassified, such alignment further degrades robustness \cite{0001ZY0MG20, dong2023enemy}. 

Following~\cite{dong2023enemy}, we construct a benign reference by taking one projected gradient step in the direction of decreasing supervised cross-entropy. In Figure~\ref{fig:3}, moving in this direction increases accuracy in the half of the samples with higher clean loss. Feature variance and input-gradient norm also decrease for this group. The update is
\begin{equation}
	\begin{aligned}
		\!\!\mathbf{\check{x}} = \projj_{\mathbb{B}(\mathbf{x}, \epsilon)} \left(  \mathbf{x} - \alpha' \cdot \operatorname{sgn} (\nabla_{\mathbf{x}}\mathcal{L}_{\text{CE}}( g_{\boldsymbol{\psi}}(f_{\boldsymbol{\theta}}\left( \mathbf{x}\right)), y))  \right),
		\label{eq:8}
	\end{aligned}
\end{equation}
where $\alpha'$ denotes the step size. Our benign refinement primarily uses one gradient descent step; the multi-step variant repeats this descent within the same input constraint. We define $\cf=f_{\boldsymbol{\theta}}(\mathbf{\check{x}})$ and compute its mean, covariance, and channel standard deviation using Eqs.~\eqref{eq:stats_1}--\eqref{eq:stats_2}. To form a benign reference distribution, we draw
\begin{equation}
\cf^\dagger=\mathcal{T}(\cf;\vs(\cf)+\vw\odot\boldsymbol{\xi}),\qquad \boldsymbol{\xi}\sim\mathcal{N}(\mathbf{0},\mI_d).
\label{eq:benign-aug}
\end{equation}
The benign branch uses random augmentation with the shared prior scale $\vw$; evolutionary selection and memory operate on the adversarial branch. The fixed reference $\cf$ defines fitness in Eq.~\eqref{eq:evo-fitness}, while the stochastic reference $\cf^\dagger$ is used for network training. For the top-$K_t$ ($K_t\le K$) elites, let $\tilde w_k=w_k/\sum_{j=1}^{K_t}w_j$. The Distribution-To-Distribution Prediction Alignment (D2D-PA) loss is
\begin{equation}
\begin{aligned}
\mathcal{L}_{\text{D2D-PA}}={}&\mathcal{L}_{\text{CE}}\big(g_{\boldsymbol{\psi}}(f_{\boldsymbol{\theta}}(\mathbf{x})),y\big)\\
&+\beta\sum_{k=1}^{K_t}\tilde w_k\KL\big(\pi_{\boldsymbol{\psi}}(\cf^\dagger)\,\|\,\pi_{\boldsymbol{\psi}}(\hf(\vu_{k:N}))\big).
\label{eq:9}
\end{aligned}
\end{equation}
This adversarial optimization can be regarded as a distributional variant of the original TRADES \cite{zhang2019theoretically} (Eq. (\ref{eq:1})). In comparison to TRADES, we consider an uncertainty-aware alignment approach between the adversarial distribution and its benignly refined distribution at the prediction level, which mitigates the potential overfitting to single adversary for each clean sample and the inductive bias induced by the local invariance of the fixed alignment reference. Overall, such an uncertainty alignment extends the traditional point-to-point prediction matching strategy to a distribution-to-distribution one to capture more generalizable and robust behavior.

Distribution-To-Distribution Statistics Alignment (D2D-SA) matches the channel statistics of the benign reference and the adversarial anchor. Their means and covariances define the Gaussian moment summaries $\mathcal{\check{N}}_{D}(\boldsymbol{\check{\mu}}, \boldsymbol{\check{\Sigma}})$ and $\mathcal{\hat{N}}_{D}(\boldsymbol{\hat{\mu}}, \boldsymbol{\hat{\Sigma}})$, respectively. The statistics-alignment loss is the KL divergence between these summaries,
{
\begin{align}
	&\mathcal{L}_\text{D2D-SA} = \mathcal{L}_{\text{KL}}\left(\mathcal{\check{N}}_{D}(\boldsymbol{\check{\mu}}, \boldsymbol{\check{\Sigma}}) \| \mathcal{\hat{N}}_{D}(\boldsymbol{\hat{\mu}}, \boldsymbol{\hat{\Sigma}}) \right) \label{eq:10} \\
	& \!\!\!=\! \frac{1}{2} \!\left(\! \operatorname{tr}\left( \boldsymbol{\hat{\Sigma}}^{-1} \boldsymbol{\check{\Sigma}} \right) \!-\! D \!+\! \left(\boldsymbol{\hat{\mu}} \!-\! \boldsymbol{\check{\mu}}\right)^{\top}\! \boldsymbol{\hat{\Sigma}}^{-1}\!\! \left(\boldsymbol{\hat{\mu}} \!-\! \boldsymbol{\check{\mu}}\right) \!+\! \operatorname{ln}\left(\!\frac{\operatorname{det}\boldsymbol{\hat{\Sigma}}}{\operatorname{det}\boldsymbol{\check{\Sigma}}} \!\right)  \!\right)\!,\!\! \nonumber
\end{align}
$\!\!\!$where $\operatorname{tr}(\cdot)$ and $\operatorname{det}(\cdot)$ are used to compute the trace and the determinant for a square matrix, respectively. The primary difference between our prediction alignment (Eq. (\ref{eq:9})) and our statistics alignment (Eq. (\ref{eq:10})) lies in the uncertainty\textemdash we incorporate the uncertainty estimation for the augmented features to capture unforeseen adversarial behavior during the distributional prediction alignment. The feature-level statistics matching acts on the unaugmented anchor $\hf$ and benign reference $\cf$. Thus the elites enter D2D-PA, while D2D-SA regularizes the reference--anchor statistics.

\subsection{Introspective Gradient Matching}

Figure \ref{fig:1_3} shows a near-linear relation between attack strength (perturbation radius) and the increase in gradient norm from clean inputs; larger radii also coincide with greater robustness degradation (Figure \ref{fig:3}). Motivated by this, we align the gradients of adversarial examples to those of their benignly refined counterparts to mitigate the degradation. While gradient alignment has been used for robustness transfer from large teachers to lightweight students, our \emph{Introspective Gradient Matching} (IGM) operates within a single model trained from scratch, without external teachers. Concretely, we match the gradient of the classification loss with respect to adversarial inputs to that of their benignly refined versions, formulated as:

\begin{equation}
\begin{aligned}
    \!\!\!\mathcal{L}_\text{IGM} \!=\! \left\| \nabla_{\!\mathbf{\check{x}}}\mathcal{L}_{\text{CE}}( g_{\boldsymbol{\psi}}(f_{\boldsymbol{\theta}}\!\left( \mathbf{\check{x}}\right)), y) \!-\! \nabla_{\!\mathbf{\hat{x}}}\mathcal{L}_{\text{CE}}( g_{\boldsymbol{\psi}}(f_{\boldsymbol{\theta}}\!\left( \mathbf{\hat{x}}\right)), y) \right\|_2\!.\!\!\!\!
    \label{eq:11}
\end{aligned}
\end{equation}
Note that the alignment reference is the benignly refined sample instead of the legitimate one, since the former exhibits better properties on both adversarial robustness and the gradient norm, which contrasts it to existing gradient penalties-based robust learning works \cite{moosavi2019robustness}. We also consider both the directions and values of two gradients inside the $\ell_{\infty}$-norm hypersphere around the given inputs for a better robust regularization during adversarial training. Thus, by aligning gradients, the model is guided to learn invariant features against adversarial perturbations, essentially focusing on the underlying data distribution rather than a certain type of noise.

\begin{algorithm}[tb]
\setstretch{0.3}
\caption{\methodname{} with Evolutionary Feature Search}
\label{alg:1}
\begin{algorithmic}[1]
\Statex {\bfseries Input:}
Dataset $\mathcal{D}$; DNN classifier $(f_{\boldsymbol{\theta}}, g_{\boldsymbol{\psi}})$; attack settings $\epsilon, \alpha, n, \ell_{\mathrm{atk}}$; benign step $\alpha'$; loss weights $\beta, \lambda_1, \lambda_2$; learning rate $\varrho$; evolutionary settings in Section~\ref{sec:AUM}.
\State Initialize network $(f_{\boldsymbol{\theta}}, g_{\boldsymbol{\psi}})$ and strategy memory $\mathcal{M}[i] \gets (\mathbf{0}, \mathbf{1}, 1, \mathbf{0})$ for each instance $i$.
\While{not at the end of adversarial training}
  \State Sample a training mini-batch $\mathcal{B} = \{(\mathbf{x}_i, y_i)\}$ from $\mathcal{D}$.
  \State \algbox{1}{Generate PGD anchors and trajectories by Eq.~(\ref{eq:2}); conduct refinement for benign references (Eq.~(\ref{eq:8})).}
  \For{each instance $i \in \mathcal{B}$ (vectorized)}
    \State \algbox{2}{Compute current anchor, style statistics, prior scale $\vw_i$, and basis $\mB_i$ via Eqs.~(\ref{eq:5})--(\ref{eq:6}).}
    \State Load stored evolution strategy $\phi_i \gets \mathcal{M}[i]$.
    \For{$t = 1, \ldots, G$}
      \State \algbox{3}{Sample and repair $N$ candidate offspring via Eq.~(\ref{eq:evo-sample}); evaluate fitness (Eq.~(\ref{eq:evo-fitness})) without gradients or BN updates.}
      \State \algbox{3}{Rank offspring and update $\phi_i$ with top-$K$ parents via Eqs.~(\ref{eq:evo-mean})--(\ref{eq:evo-step}).}
    \EndFor
    \State \algbox{2}{Store $\mathcal{M}[i] \gets \phi_i$ and retain the last generation's top-$K_t$ candidate offsets.}
    \State \algbox{2}{Decode retained features with gradients via Eq.~(\ref{eq:7}); sample benign $\cf_i^\dagger$ via Eq.~(\ref{eq:benign-aug}).}
  \EndFor
  \State \algbox{1}{Average loss terms in Eqs.~(\ref{eq:9})--(\ref{eq:11}) over mini-batch $\mathcal{B}$ and aggregate total objective $\mathcal{L}$ via Eq.~(\ref{eq:12}).}
  \State \algbox{1}{Update network parameters: $(\boldsymbol{\theta}, \boldsymbol{\psi}) \gets (\boldsymbol{\theta}, \boldsymbol{\psi}) - \varrho \nabla_{(\boldsymbol{\theta}, \boldsymbol{\psi})}\mathcal{L}$, holding sampled offsets and strategy parameters $\phi_i$ fixed.}
\EndWhile \\
\Return Trained network $(f_{\boldsymbol{\theta}}, g_{\boldsymbol{\psi}})$, retaining the primary BN at inference.
\end{algorithmic}
\end{algorithm}

\subsection{Theoretical Analysis of Evolutionary Selection}
\label{sec:theory-uadat}

We analyze one instance at one visit, conditional on the current network, the benign reference $\cf$, and the strategy $\phi$ that generated the population. The candidate evaluator is deterministic, with fixed normalization statistics. Write $\vu\sim q_\phi$ for a draw from the Gaussian proposal and $\tF(\vu)=F(\Pi_\tau(\vu))$ for its repaired fitness. The analysis concerns the fitness surrogate evaluated against $\cf$. The D2D-PA training loss in Eq.~\eqref{eq:9} uses the stochastic reference $\cf^\dagger$.

\begin{proposition}[Bounded deviation from the prior]
\label{prop:evo-kl}
Let $p_0=\mathcal{N}(\mathbf{0},\mI_d)$ and assume $dc_{\max}\ge1$. Every maintained Gaussian proposal, before repair, satisfies $\KL(q_\phi\|p_0)\le\bar\rho$, where
\begin{equation}
\bar\rho=\frac d2\left(\eta_{\max}^2c_{\max}+\tau^2-1-\ln(\gamma\eta_{\min}^2c_{\min})\right).
\label{eq:evo-kl}
\end{equation}
\end{proposition}
\begin{proof}
The Gaussian KL equals $\tfrac12(\tr\mC+\|\vm\|_2^2-d-\ln\det\mC)$. The mean repair gives $\|\vm\|_2^2\le\tau^2d$. Since $\mB^\top\mB=\mI_r$, Eq.~\eqref{eq:evo-cov} gives $\tr\mC\le\eta_{\max}^2dc_{\max}$ and $\mC\succeq\gamma\eta_{\min}^2c_{\min}\mI_d$, proving the bound.
\end{proof}

For a tail fraction $\upsilon\in(0,1]$ (distinct from the training weight $\beta$), define the likelihood-ratio set $\mathcal{A}_\upsilon(q)=\{q'\ll q:\mathrm{d}q'/\mathrm{d}q\le1/\upsilon\}$ and the upper-tail risk $\mathrm{CVaR}_\upsilon(\tF;q)=\sup_{q'\in\mathcal{A}_\upsilon(q)}\E_{q'}\tF$~\cite{rockafellar2002cvar}. Let $J(q)=\E_q\tF$ and $R(\rho)=\sup_{q':\KL(q'\|p_0)\le\rho}\E_{q'}\tF$. For normalized nonincreasing elite weights, set
\begin{equation}
\begin{aligned}
\widehat H&=\sum_{k=1}^{K_t}\tilde w_k\tF_{k:N},\\
S_w(q)&=\sum_{j=1}^{K_t}a_j\mathrm{CVaR}_{j/N}(\tF;q),
\end{aligned}
\label{eq:evo-risk}
\end{equation}
where $a_j=j(\tilde w_j-\tilde w_{j+1})\ge0$, $\tilde w_{K_t+1}=0$, and $\sum_j a_j=1$. The repaired search region is compact, so continuity of the network gives $0\le\tF\le F_{\max}<\infty$.

\begin{theorem}[Elite risk and a distributionally robust upper bound]
\label{thm:evo-risk}
For $N$ independent draws from the fixed proposal $q_\phi$, let $\rho_\upsilon=\ln(1/\upsilon)+(\bar\rho+1)/\upsilon$. Then
\begin{equation}
J(q_\phi)\le\E[\widehat H]\le S_w(q_\phi)\le R(\rho_{1/N}).
\label{eq:evo-sandwich}
\end{equation}
For any $\delta\in(0,1)$, with probability at least $1-\delta$,
\begin{equation}
|\widehat H-S_w(q_\phi)|\le e_N,\quad
 e_N=N\tilde w_1F_{\max}\sqrt{\frac{\ln(2/\delta)}{2N}}.
\label{eq:evo-concentration}
\end{equation}
\end{theorem}
\begin{proof}
Let $\widehat C_j$ be the average of the largest $j$ sampled fitnesses. Abel summation gives $\widehat H=\sum_j a_j\widehat C_j$. Each $\widehat C_j$ dominates the full sample mean; the variational formula $\mathrm{CVaR}_\upsilon=\inf_t\{t+\upsilon^{-1}\E(\tF-t)_+\}$ gives $\E\widehat C_j\le\mathrm{CVaR}_{j/N}$. For $q'\in\mathcal{A}_\upsilon(q_\phi)$, put $r=\mathrm{d}q'/\mathrm{d}q_\phi$ and $\ell_0=\ln(\mathrm{d}q_\phi/\mathrm{d}p_0)$. Since $\E_{q_\phi}(\ell_0)_-\le1$, Proposition~\ref{prop:evo-kl} yields $\KL(q'\|p_0)=\E[r\ln r+r\ell_0]\le\ln(1/\upsilon)+(\bar\rho+1)/\upsilon$. This bounds each CVaR term by $R(\rho_{j/N})\le R(\rho_{1/N})$. Finally, the Dvoretzky--Kiefer--Wolfowitz inequality~\cite{massart1990dkw} bounds the empirical survival-function error by $\sqrt{\ln(2/\delta)/(2N)}$. Integrating over $[0,F_{\max}]$ gives a factor $N/j$ for each tail average; summing with $\sum_j a_jN/j=N\tilde w_1$ proves Eq.~\eqref{eq:evo-concentration}.
\end{proof}

\textbf{Conditional interpretation for latent adversaries.} If the benign reference is confidently correct, $[\pi_{\boldsymbol{\psi}}(\cf)]_y\ge1-\vartheta$ for $\vartheta\in[0,1/2)$, let $c_\vartheta=\ln2-h(\vartheta)>0$, where $h$ is binary entropy. Misclassification of $\hf(\Pi_\tau(\vu))$ implies that its correct-class probability is at most $1/2$, and binary KL data processing gives $\tF(\vu)\ge c_\vartheta$~\cite{cover2006elements}. Consequently, on the event in Eq.~\eqref{eq:evo-concentration},
\begin{equation}
\sup_{q'\in\mathcal{A}_{K_t/N}(q_\phi)}\Pr_{q'}(\text{misclassification})
\le\min\left\{1,\frac{\widehat H+e_N}{c_\vartheta}\right\}.
\label{eq:evo-certificate}
\end{equation}
The inequality follows from $\mathrm{CVaR}_{K_t/N}\le S_w$. Under the confidence condition on the fixed benign reference, Eq.~\eqref{eq:evo-certificate} bounds the misclassification probability of decoded feature candidates for distributions $q'\in\mathcal{A}_{K_t/N}(q_\phi)$.

\subsection{Objective Function}
We here formalize the overall loss during adversarial training by combining the distribution-to-distribution prediction alignment loss $\mathcal{L}_\text{D2D-PA}$, the statistics alignment loss $\mathcal{L}_\text{D2D-SA}$, and the introspective gradient matching loss $\mathcal{L}_\text{IGM}$ as below:
\begin{equation}
	\begin{aligned}
		\mathcal{L} = \mathcal{L}_\text{D2D-PA} + \lambda_1 \cdot \mathcal{L}_\text{D2D-SA} + \lambda_2 \cdot \mathcal{L}_\text{IGM}
		\label{eq:12}
	\end{aligned}
\end{equation}
where $\lambda_1>0$ and $\lambda_2>0$ denote the loss weighting hyper-parameters. These per-instance losses are averaged over the mini-batch. During network optimization, sampled offsets and strategy parameters are held fixed, while gradients propagate through the decoded features and network. The pseudocode of our proposed \methodname{} method is provided in Algorithm \ref{alg:1}. During the inference stage, we can directly adopt the adversarially trained model to classify both clean and adversarial examples.

Considering the inherent domain shift between the adversarial distribution and its benignly refined counterpart \cite{xie2020adversarial}, we introduce separate Batch Normalization (BN) layers for these two branches, thereby aiding in the disentanglement of the mixed distributions. Specifically, the primary BN layer is designated for processing clean images (or benignly refined samples). Conversely, an auxiliary BN layer is designed for handling adversarial samples. During the inference stage, we keep the primary BN layer only, which means that our method has no differences compared to others at the test time.

\section{Experiments}
In this section, we measure clean and robust accuracy on four image-classification benchmarks. Table~\ref{tab:1} reports PGD-20, Carlini--Wagner, and AutoAttack accuracy for ResNet-18. The single-step comparison in Table~\ref{tab:5} uses PGD-50 with ten restarts. We also evaluate larger backbones and training with auxiliary synthetic data.

\begin{table*}[!t]
\centering
\vspace{-3mm}
\caption{Comparison of \methodname{} using ResNet-18 with other adversarial training approaches on CIFAR-10, CIFAR-100, and SVHN. The $\ell_{\infty}$-norm perturbations are restricted in the radius $\epsilon = 8/255$. We report both clean and robust accuracy (\%).}
\vspace{-2mm}
\renewcommand{\arraystretch}{0.5}
\resizebox{\linewidth}{!}{
\begin{tabular}{ccccccccccccc}
    \toprule
    \multirow{2}{*}{Method} &\multicolumn{4}{c}{CIFAR-10}&\multicolumn{4}{c}{CIFAR-100}&\multicolumn{4}{c}{SVHN}\\
    \cmidrule(r){2-5} \cmidrule(r){6-9} \cmidrule(r){10-13}
    & Clean & PGD & CW & AA & Clean & PGD & CW & AA & Clean & PGD & CW & AA\\
    \midrule
    PGD-AT \cite{MadryMSTV18} & 83.80  & 51.40 & 50.17  & 47.68 & 57.39  & 28.36 & 26.29  & 23.18 & 92.46  & 50.55 & 50.40  & 46.07\\
    TRADES \cite{zhang2019theoretically} & 82.45  & 52.21 & 50.29  & 48.88 & 54.36  & 27.49 & 24.19  & 23.14 & 90.63  & 58.10 & 55.13  & 52.62\\
    MART \cite{0001ZY0MG20} & 82.20 & 53.94 & 50.43  & 48.04 & 54.78  & 28.79 & 26.15  & 24.58 & 89.88  & 58.48 & 52.48 & 48.44\\
    HAT \cite{RadeM22} & 84.86  & 52.04 & 50.33 & 48.85 & 58.73  & 27.92 & 24.60  & 23.34 & 92.06 & 57.35 & 54.77  & 52.06\\
    APT \cite{10177878} & 83.47 & 54.24 & 51.92 & 50.06 & 58.27 & 31.05 & 27.49 & 26.02 & 92.32 & 57.80 & 53.69 & 50.26 \\
    UIAT \cite{dong2023enemy} & 85.01  & 54.63 & 51.10  & 49.11 & 59.55  & 30.81 & 28.05  & 25.73 & 93.28  & 58.18 & 55.49  & 52.45\\
    ARoW \cite{yang2023improving} & 83.46 & 54.07 & 51.24 & 50.29 & 59.98 & 31.10 & 27.61 & 25.96 & 93.15 & 58.29 & 55.68 & 52.19 \\
    DHAT \cite{Zhang_2025_ICCV} & 85.59 & 54.46 & 52.08 & 50.34 & 60.74 & 31.48 & 28.47 & 26.37 & 93.39 & 58.90 & 55.87 & 52.62 \\
    RAAT \cite{Wang_2026_CVPR} & 84.67 & 54.29 & 52.37 & 49.88 & 60.02 & 28.94 & 26.73 & 24.81 & 92.85 & 57.98 & 54.62 & 50.87 \\
    \rowcolor{LightCyan}\textbf{EvoUAD-AT} & \textbf{86.83} & \textbf{55.78} & \textbf{53.35} & \textbf{52.02} & \textbf{62.25} & \textbf{33.18} & \textbf{28.76} & \textbf{27.39} & \textbf{93.78} & \textbf{59.96} & \textbf{56.54} & \textbf{53.51} \\
    \bottomrule
\end{tabular}
}
\vspace{-4mm}
\label{tab:1}
\end{table*}

\subsection{Experimental Setups}
\noindent \textbf{Datasets.}
Following the evaluation criteria on RobustBench \cite{CroceASDFCM021}, we conduct experiments across four standard datasets: CIFAR-10, CIFAR-100 \cite{krizhevsky2009learning}, SVHN \cite{netzer2011reading}, and ImageNet \cite{deng2009imagenet}. The CIFAR-10 dataset comprises 60,000 color images with a resolution of 32 $\times$ 32 pixels, distributed across 10 distinct classes. CIFAR-100 mirrors the same structure of CIFAR-10 but expands the classification complexity with 100 categories. The SVHN dataset contains 73,257 training samples and 26,032 test samples for digit recognition in the context of street view imagery. ImageNet-1K is a large-scale image classification benchmark comprising 1,000 classes, which is utilized for robustness evaluation in the context of diverse real-world images. Furthermore, for adversarial training with auxiliary data, we incorporate one million synthetic images for CIFAR-10/100, generated by the Denoising Diffusion Probabilistic Model (DDPM) \cite{ho2020denoising} in line with the experimental protocols in prior works \cite{rebuffi2021fixing, RadeM22, dong2023enemy}.

\looseness=-1
\noindent \textbf{Implementation details.}
In line with the protocols of previous adversarial training works \cite{zhang2019theoretically, RadeM22, dong2023enemy} and RobustBench \cite{CroceASDFCM021}, we adopt ResNet-18/50, Pre-activation ResNet-18 (PRN-18), Wide-ResNet-28-10 (WRN-28-10), and WRN-70-16 network architectures. For experiments that do not involve additional training data, we employ a batch size of $128$ with $100$ training epochs for CIFAR-10/100 and $50$ epochs for SVHN. The optimization process is based on Stochastic Gradient Descent (SGD) with a Nesterov momentum factor of $0.9$, a weight decay of $5 \times 10^{-4}$, and a cyclic learning rate scheduling strategy that peaks at $0.1$. In the context of training with auxiliary synthetic data for CIFAR-10/100 \cite{rebuffi2021fixing, wang2023better}, we conduct adversarial training for $800$ CIFAR-equivalent epochs (the same amount of training samples encountered in a standard CIFAR-10/100 epoch) with a batch size $1024$ and a learning rate of $0.4$. The composition of training batches exhibits an original-to-synthetic image ratio of $0.3$ for CIFAR-10, which implies the inclusion of $7$ synthetic images for every $3$ original images, while this ratio is adjusted to $0.4$ for CIFAR-100. For adversary generation during the training stage, we resort to the PGD method \cite{MadryMSTV18} applied to cross-entropy loss with 10 iterative steps (step size $\alpha=2/255$ for CIFAR-10/100 and $\alpha=1/255$ for SVHN). The step size for the benign refinement is set to $\alpha'=8/255$. We primarily focus on the $\ell_{\infty}$-norm threat model with the perturbation radius $\epsilon=8/255$. The trade-off factor is set to $\beta=4.0$ for CIFAR-10/100 and $\beta=3.0$ for SVHN. The evolutionary configuration is described in Section~\ref{sec:AUM}: the current trajectory determines the prior scale and guidance, and an instance-indexed memory stores the strategy parameters across epochs. We determine the loss weighting factors $\lambda_1=1.0$ and $\lambda_2=0.05$ through cross-validation on CIFAR-10 and consistently apply them across all datasets.

\subsection{Experimental Results}

\noindent\textbf{Performance of \methodname{}.}
Table~\ref{tab:1} compares ResNet-18 models on CIFAR-10, CIFAR-100, and SVHN. We report clean accuracy and robust accuracy under PGD with 20 steps of size $\alpha=2/255$~\cite{MadryMSTV18}, the Carlini--Wagner attack (CW)~\cite{carlini2017towards}, and AutoAttack (AA)~\cite{croce2020reliable}. AutoAttack combines three white-box attacks with one black-box attack. On each dataset, \methodname{} has the highest clean, PGD, CW, and AA accuracies among the methods in the table.

\begin{table}[!t]
\centering
\renewcommand{\arraystretch}{0.5}
\vspace{1.24mm}
\caption{Robust accuracy (\%) on ImageNet using ResNet-50 under the $\ell_\infty$ threat model with $\epsilon=4/255$.}
\vspace{-2mm}
\resizebox{0.8\linewidth}{!}{
\begin{tabular}{ccccc}
\toprule
Method & Clean & PGD & CW & AA \\
\midrule
TRADES \cite{zhang2019theoretically} & 62.65 & 38.46 & 36.15 & 34.99 \\
HAT \cite{RadeM22} & 64.23 & 38.98 & 36.53 & 35.24 \\
UIAT \cite{dong2023enemy} & 64.90 & 40.27 & 37.39 & 36.08 \\
ARoW \cite{yang2023improving} & 64.72 & 40.18 & 37.70 & 36.25 \\
DHAT \cite{Zhang_2025_ICCV} & 65.53 & 40.54 & 37.95 & 36.69 \\
RAAT \cite{Wang_2026_CVPR} & 64.38 & 39.37 & 37.41 & 35.33 \\
\rowcolor{LightCyan}\textbf{EvoUAD-AT} & \textbf{66.41} & \textbf{41.75} & \textbf{38.59} & \textbf{37.40} \\
\bottomrule

\end{tabular}
}
\vspace{-1mm}
\label{tab:2}
\end{table}

\noindent\textbf{Adversarial training on ImageNet.}
We further investigate the scalability and generalization ability of \methodname{} on ImageNet-1K, a large-scale image classification benchmark with 1,000 classes. As shown in Table \ref{tab:2}, we compare \methodname{} with existing adversarial training methods using ResNet-50, evaluating both clean accuracy and adversarial robustness. This experiment examines whether the proposed uncertainty-aware distributional modeling framework remains effective when scaling from smaller image classification benchmarks to a more challenging large-scale setting.

\begin{table}[!t]
\centering
\vspace{-1mm}
\caption{Adversarial robustness under different attack configurations using ResNet-18 on CIFAR-10. We present clean accuracy (\%) and (Auto-Attack) robust accuracy (\%).}
\vspace{-2mm}
\renewcommand{\arraystretch}{0.5}
\resizebox{0.9\linewidth}{!}{
\begin{tabular}{ccccccc}
\toprule
\multirow{2}{*}{Method} &\multicolumn{2}{c}{$\epsilon=10/255$}&\multicolumn{2}{c}{$\epsilon=12/255$}&\multicolumn{2}{c}{$\epsilon=16/255$} \\
\cmidrule(r){2-3} \cmidrule(r){4-5} \cmidrule(r){6-7}
& Clean & Robust & Clean & Robust & Clean & Robust \\
\midrule
TRADES \cite{zhang2019theoretically} & 82.28 & 38.55 & 79.37 & 31.84 & 74.89 & 18.70 \\
HAT \cite{RadeM22} & 81.94 & 40.12 & 79.43 & 33.28 & 74.45 & 19.42 \\
APT \cite{10177878} & 81.43 & 40.20 & 79.08 & 33.39 & 73.71 & 20.16 \\
UIAT \cite{dong2023enemy} & 82.79 & 40.61 & 79.50 & 34.32 & 74.29 & 21.82 \\
ARoW \cite{yang2023improving} & 81.43 & 40.36  & 79.24  & 33.81  & 74.11  & 20.97 \\
DHAT \cite{Zhang_2025_ICCV} & 82.80 & 40.68 & 80.01 & 34.59 & 74.80 & 21.63 \\
RAAT \cite{Wang_2026_CVPR} & 82.49 & 40.09 & 79.92 & 33.28 & 74.84 & 20.05 \\
\rowcolor{LightCyan}\textbf{EvoUAD-AT} & \textbf{83.66} & \textbf{41.43} & \textbf{80.51} & \textbf{35.90} & \textbf{75.78} & \textbf{22.59} \\
\bottomrule
\end{tabular}
}
\vspace{-3mm}
\label{tab:3}
\end{table}

\noindent\textbf{Adversarial training with larger perturbation radius.}
In addition to the standard evaluation setting on adversarial examples with the perturbation radius $\epsilon=8/255$, we consider more difficult adversary configurations when confronted with adversaries with larger perturbation radii. Specifically, we conduct adversarial training based on various large $\ell_{\infty}$-norm perturbation radii: 10/255, 12/255, and 16/255, respectively. Afterward, we evaluate the robustness on adversarial examples with the corresponding perturbation radius for consistency, as shown in Table \ref{tab:3}. As observed, our \methodname{} can achieve an overall better classification performance on both clean and adversarial examples in comparison with other approaches. This finding also reveals that our method not only enhances robustness against progressively intensive adversarial attacks but also maintains better natural performance in diverse settings. In other words, our \methodname{} method can achieve enhanced robustness against stronger adversarial attacks via conducting adversarial training with larger perturbation radii. 

\begin{table}[!t]
\centering
\vspace{-2mm}
\renewcommand{\arraystretch}{0.5}
\caption{$\mathcal{L}_2$ robust acc. (\%) on CIFAR-10 with ResNet-18.}
\vspace{-2mm}
\resizebox{0.8\linewidth}{!}{
\begin{tabular}{ccccc}
\toprule
Method & Clean & PGD & CW & AA \\
\midrule
PGD-AT \cite{MadryMSTV18} & 86.99 & 73.40 & 69.56 & 67.11 \\
TRADES \cite{zhang2019theoretically} & 87.38 & 74.93 & 70.48 & 68.31 \\
MART \cite{0001ZY0MG20} & 87.02 & 74.89 & 69.79 & 67.86 \\
HAT \cite{RadeM22} & 87.95 & 75.10 & 70.83 & 68.92 \\
APT \cite{10177878} & 87.79 & 75.04 & 70.62 & 68.59 \\
UIAT \cite{dong2023enemy} & 88.14 & 75.28 & 70.91 & 68.89 \\
ARoW \cite{yang2023improving} & 88.07 & 75.16 & 70.95 & 68.96 \\
DHAT \cite{Zhang_2025_ICCV} & 88.25 & 75.40 & 71.09 & 69.02 \\
RAAT \cite{Wang_2026_CVPR} & 88.04 & 74.29 & 70.85 & 68.63 \\
\rowcolor{LightCyan}\textbf{EvoUAD-AT} & \textbf{89.63} & \textbf{76.58} & \textbf{72.30} & \textbf{70.02} \\
\bottomrule

\end{tabular}
}
\vspace{-3mm}
\label{tab:supp_L2}
\end{table}

\noindent\textbf{Robustness against $\mathcal{L}_{2}$-norm adversaries.}
To provide a comprehensive robustness evaluation, we here follow the setting from RobustBench \cite{CroceASDFCM021} to incorporate the experimental validation of adversarial robustness under the $\mathcal{L}_2$-norm threat model in comparisons with other adversarial training works in Table \ref{tab:supp_L2}. In the case of $\mathcal{L}_2$-norm adversarial perturbations, we use the following adversary setting during the training stage: perturbation radius $\epsilon=128/255$, and the step size $\alpha = 15/255$ with $10$ steps. We report both natural performance and robust accuracy against these $\mathcal{L}_2$-norm adversarial examples using the ResNet-18 network architecture. We can observe that our \methodname{} still achieves the best performance in terms of clean and robust accuracy.

\begin{table}[!t]
\centering
\renewcommand{\arraystretch}{0.5}
\caption{Robust acc. (\%) on CIFAR-10-C with ResNet-18.}
\vspace{-2mm}
\resizebox{0.6\linewidth}{!}{
\begin{tabular}{ccc}
\toprule
Method & Clean & Corruption \\
\midrule
PGD-AT \cite{MadryMSTV18} & 91.45 & 79.37 \\
TRADES \cite{zhang2019theoretically} & 92.35 & 80.54 \\
MART \cite{0001ZY0MG20} & 91.74 & 80.49 \\
HAT \cite{RadeM22} & 92.76 & 81.46 \\
APT \cite{10177878} & 92.02 & 80.15 \\
UIAT \cite{dong2023enemy} & 92.97 & 82.38 \\
ARoW \cite{yang2023improving} & 92.19 & 82.75 \\
DHAT \cite{Zhang_2025_ICCV} & 92.48 & 82.38\\
RAAT \cite{Wang_2026_CVPR} & 92.90 & 81.46\\
\rowcolor{LightCyan}\textbf{EvoUAD-AT} & \textbf{94.16} & \textbf{84.21} \\
\bottomrule

\end{tabular}
}
\vspace{-3mm}
\label{tab:supp_Corruption}
\end{table}

\noindent\textbf{Robustness against corruption.}
To explore the generalization ability of our method, we conduct robustness evaluations on common corruptions from the CIFAR-10-C benchmark \cite{HendrycksD19}. The $\mathcal{L}_\infty$ perturbation radius $\epsilon$ of all the adversarial training methods is set as $1/255$ with the step size $\alpha=0.25/4$. As shown in Table \ref{tab:supp_Corruption}, we report both the clean accuracy and also corruption accuracy using ResNet-18. Note that all the adversarial training methods are trained on the original CIFAR-10 dataset and subsequently evaluated on CIFAR-10-C. We can easily observe that our \methodname{} approach also obtains the state-of-the-art performance in terms of clean and corruption accuracy. In addition, we improve our baseline TRADES \cite{zhang2019theoretically} by a large margin, further demonstrating the generalization ability of our proposed method.

\noindent\textbf{Adversarial training with auxiliary data.}
Previous adversarial training works have demonstrated the efficacy of augmenting the training set with additional synthetic data \cite{rebuffi2021fixing, RadeM22, wang2023better, dong2023enemy}. Here, we investigate the generalization ability of our \methodname{} method when supplemented with auxiliary generated data using larger network architectures. Specifically, we report both natural performance and adversarial robustness results of the adversarially trained models built upon the composition of original CIFAR-10/100 datasets and their synthetic counterparts generated by DDPM and Elucidating Diffusion Model (EDM) following \cite{rebuffi2021fixing, wang2023better}, as shown in Table \ref{tab:4}. Note that the CutMix augmentation technique is omitted during the training process in consistent with \cite{RadeM22}. The results underscore the efficacy of our method, which not only enhances robust accuracy but also preserves or even improves natural performance in comparison to established benchmarks.

\begin{table}[!t]
\centering
\vspace{-2mm}
\caption{Comparison of adversarial training on CIFAR-10/CIFAR-100 with 1M auxiliary synthetic data. We report clean accuracy (\%) and (Auto-Attack) robust accuracy (\%).}
\vspace{-2mm}
\renewcommand{\arraystretch}{0.7}
\resizebox{0.9\linewidth}{!}{
\begin{tabular}{ccccccc}
\toprule
\multirow{2}{*}{Architecture} & \multirow{2}{*}{\makecell{Generative \\ Model}} & \multirow{2}{*}{Method} & \multicolumn{2}{c}{CIFAR-10} & \multicolumn{2}{c}{CIFAR-100} \\
\cmidrule(r){4-5} \cmidrule(r){6-7}
&&& Clean & Robust & Clean & Robust\\
\midrule
\multirow{7}{*}{PRN-18} & \multirow{7}{*}{DDPM} & Rebuffi \textit{et al.} \cite{rebuffi2021fixing}& 83.53 & 56.66 & 56.87 & 28.50 \\
&& HAT \cite{RadeM22} & 86.86 & 57.09 & 61.50 & 28.88 \\
&& APT \cite{10177878} & 85.21 & 56.33 & 59.17 & 28.24 \\
&& UIAT \cite{dong2023enemy} & 87.34 & 58.46 & 62.20 & 29.40 \\
&& ARoW \cite{yang2023improving} & 86.92 & 58.11 & 61.25 & 29.07 \\
&& RAAT \cite{Wang_2026_CVPR} & 87.09 & 57.25 & 61.09 & 28.46 \\
&& \cellcolor{LightCyan}\textbf{EvoUAD-AT} & \cellcolor{LightCyan}\textbf{87.74} & \cellcolor{LightCyan}\textbf{59.48} & \cellcolor{LightCyan}\textbf{62.33} & \cellcolor{LightCyan}\textbf{30.08} \\
\midrule
\multirow{9}{*}{WRN-28} & \multirow{7}{*}{DDPM} & Rebuffi \textit{et al.} \cite{rebuffi2021fixing}& 85.97 & 60.73 & 59.18 & 30.81 \\
&& HAT \cite{RadeM22} & 88.16 & 60.97 & 62.21 & 31.16 \\
&& APT \cite{10177878} & 86.48 & 60.18 & 61.93 & 30.72 \\
&& UIAT \cite{dong2023enemy} & 88.93 & 61.32 & 63.26 & 31.18 \\
&& ARoW \cite{yang2023improving} & 88.50 & 61.14 & 62.78 & 30.93 \\
&& RAAT \cite{Wang_2026_CVPR} & 88.63 & 60.87 & 62.16 & 30.35 \\
&& \cellcolor{LightCyan}\textbf{EvoUAD-AT} & \cellcolor{LightCyan}\textbf{89.14} & \cellcolor{LightCyan}\textbf{62.08} & \cellcolor{LightCyan}\textbf{63.83} & \cellcolor{LightCyan}\textbf{31.52} \\
\cmidrule(r){2-7}
&\multirow{2}{*}{EDM} & Wang \textit{et al.} \cite{wang2023better}& 91.12 & 63.35 & 68.06 & 35.65 \\
&& \cellcolor{LightCyan}\textbf{EvoUAD-AT} & \cellcolor{LightCyan}\textbf{91.59} & \cellcolor{LightCyan}\textbf{64.27} & \cellcolor{LightCyan}\textbf{68.45} & \cellcolor{LightCyan}\textbf{36.27} \\
\midrule
\multirow{2}{*}{WRN-70} & \multirow{2}{*}{EDM} & Wang \textit{et al.} \cite{wang2023better}& 91.98 & 65.54 & 70.21 & 38.69 \\
&& \cellcolor{LightCyan}\textbf{EvoUAD-AT} & \cellcolor{LightCyan}\textbf{92.34} & \cellcolor{LightCyan}\textbf{66.25} & \cellcolor{LightCyan}\textbf{70.96} & \cellcolor{LightCyan}\textbf{39.34} \\
\bottomrule

\end{tabular}
}
\vspace{-4mm}
\label{tab:4}
\end{table}

\begin{table*}[!t]
	\centering
    \vspace{-3mm}
	\caption{Extension with single-step adversarial training methods on CIFAR-10. We conduct single-step adversarial training with various perturbation radii. We report the clean accuracy (\%), robust accuracy (\%) on PGD-based adversaries with 50 steps and 10 restarts, and the average training time (seconds) for an epoch.}
	\vspace{-2mm}
	\renewcommand{\arraystretch}{0.5}
	\resizebox{\linewidth}{!}{
	\begin{tabular}{cccccccccccc}
		\toprule
		\multirow{2}{*}{Method}&\multicolumn{2}{c}{$\epsilon=4/255$}&\multicolumn{2}{c}{$\epsilon=6/255$}&\multicolumn{2}{c}{$\epsilon=8/255$}&\multicolumn{2}{c}{$\epsilon=10/255$}&\multicolumn{2}{c}{$\epsilon=12/255$}&\multirow{2}{*}{Time(s)}\\
		\cmidrule(lr){2-3} \cmidrule(lr){4-5} \cmidrule(lr){6-7} \cmidrule(lr){8-9} \cmidrule(lr){10-11}
		&Clean&Robust&Clean&Robust&Clean&Robust&Clean&Robust&Clean&Robust&\\
		\midrule
		N-FGSM \cite{de2022make} & 87.24 & 66.69 & 84.66 & 56.36 & 80.29 & 48.24 & 75.59 & 41.54 & 71.01 & 36.92 & 48.4 \\
		\rowcolor{LightCyan}N-FGSM\textbf{ + EvoUAD-AT} & 88.61 & 68.95 & 85.61 & 58.99 & 82.06 & 50.15 & 78.26 & 43.08 & 73.35 & 38.95 & 54.9 \\
		\midrule
		RS-FGSM \cite{WongRK20} & 88.18 & 66.64 & 86.72 & 55.28 & 84.07 & 46.15 & 86.32 & 0.00 & 85.55 & 0.00 & 32.4 \\
		\rowcolor{LightCyan}RS-FGSM\textbf{ + EvoUAD-AT} & 88.79 & 67.03 & 87.78 & 56.34 & 85.21 & 46.82 & 88.13 & 0.00 & 87.34 & 0.00 &  36.7 \\
		\midrule
		GradAlign \cite{andriushchenko2020understanding} & 88.00 & 65.74 & 83.85 & 55.25 & 80.17 & 46.57 & 76.46 & 39.85 & 72.38 & 34.82 & 96.0 \\
		\rowcolor{LightCyan}GradAlign\textbf{ + EvoUAD-AT} & 89.71 & 66.86 & 85.84 & 56.31 & 82.30 & 47.22 & 79.20 & 40.37 & 74.15 & 37.33 & 104.2 \\
		
		\bottomrule
	\end{tabular}
	}
	\vspace{-4mm}
	\label{tab:5}
\end{table*}

\noindent\textbf{Extension with single-step adversary generation.}
Single-step training reduces the cost of generating adversarial inputs~\cite{shafahi2019adversarial}. We integrate \methodname{} with N-FGSM~\cite{de2022make}, RS-FGSM~\cite{WongRK20}, and GradAlign~\cite{andriushchenko2020understanding}. With $n=1$, we set $r=0$ and $\gamma=1$, which leaves a diagonal sampling covariance. Benign refinement takes one descent step. We use IGM only with GradAlign, matching the input gradients of the adversarial and benignly refined samples. At $\epsilon=8/255$, adding \methodname{} to N-FGSM raises robust accuracy from $48.24\%$ to $50.15\%$ under PGD-50 with ten restarts (Table~\ref{tab:5}). Training time rises from $48.4$ to $54.9$ seconds per epoch. At $\epsilon=10/255$ and $12/255$, RS-FGSM has $0\%$ robust accuracy both alone and with \methodname{}.

\section{Analyses}

\begin{table}[!t]
	\centering
    \vspace{0mm}
	\caption{Ablation study using ResNet-18 of three components of EvoUAD-AT for accuracy (\%) on CIFAR-10. }
	\vspace{-2mm}
	\renewcommand{\arraystretch}{0.8}
	\resizebox{0.9\linewidth}{!}{
		\begin{tabular}{c|cccccc}
			\toprule
			&  EvoAUM & DAR & IGM & Clean & PGD & AA \\
			\midrule
			1 & & &  & 82.45 & 52.21 & 48.88    \\
			2 & \checkmark & & & 85.35 & 54.90 & 50.49     \\
			3 & \checkmark & \checkmark & & 86.34 & 55.18 & 50.49     \\
			4 & \checkmark & & \checkmark & 85.68 & 55.40 & 51.15     \\
			\midrule
			\rowcolor{LightCyan}5 & \checkmark & \checkmark & \checkmark & 86.83 & 55.78 & 52.02     \\
			
			\bottomrule
			\multicolumn{7}{l}{\footnotesize{EvoAUM: Evolutionary Adversarial Uncertainty Modeling.}}\\
			\multicolumn{7}{l}{\footnotesize{DAR: Distributional Alignment Refinement.}}\\
			\multicolumn{7}{l}{\footnotesize{IGM: Introspective Gradient Matching.}}
		\end{tabular}
		}
        \vspace{-2mm}
		\label{tab:6}
\end{table}

\subsection{Ablation Study}

\noindent\textbf{Impact of each module.}
Table~\ref{tab:6} isolates the contributions of \modname{} (Section~\ref{sec:AUM}), DAR (Section~\ref{sec:MethodDAR}), and IGM (Eq.~\eqref{eq:11}) on CIFAR-10. Adding \modname{} to TRADES~\cite{zhang2019theoretically} raises AutoAttack accuracy from $48.88\%$ to $50.49\%$. Adding DAR to \modname{} raises clean accuracy from $85.35\%$ to $86.34\%$, with AutoAttack accuracy at $50.49\%$ in both settings. Combining \modname{} with IGM gives $51.15\%$ AutoAttack accuracy. The full method reaches $86.83\%$ clean accuracy and $52.02\%$ AutoAttack accuracy.

\begin{table}[!t]
	\centering
	\caption{Comparison (Acc. (\%)) of diverse adversary modeling strategies using ResNet-18 on CIFAR-10.}
	\vspace{-2mm}
	\renewcommand{\arraystretch}{0.5}
	\resizebox{0.9\linewidth}{!}{
		\begin{tabular}{cccc}
			\toprule
			Modeling Strategy & Clean & PGD & AA \\
			\midrule
			Single Point (TRADES \cite{zhang2019theoretically}) & 82.45 & 52.21 & 48.88 \\
			Random Statistics ($\mathcal{N}(\mathbf{0}, \mathbf{I})$) & 81.97 & 49.83 & 43.29 \\
			Random Statistics ($\mathcal{N}(\mu(\mathbf{x}), \mathbf{I})$) & 84.40 & 50.96 & 45.62 \\
			Random Statistics ($\mathcal{N}(\mu(\mathbf{\hat{x}}), \mathbf{I})$) & 82.34 & 51.76 & 47.86 \\
			\textbf{EvoAUM} (w/ Intermediate Adversaries) & 84.18 & 53.98 & 49.03 \\
			\textbf{EvoAUM} (w/ Historical Adversaries) & 84.41 & 54.15 & 49.19 \\
			\rowcolor{LightCyan}\textbf{EvoAUM} (Ours) & 85.35 & 54.90 & 50.49 \\
			
			\bottomrule
		\end{tabular}
		}
	\vspace{-2mm}
	\label{tab:7}
\end{table}

\noindent\textbf{Performance w.r.t. adversary modeling.}
In addition to our evolutionary adversary modeling scheme, denoted as \modname{} in previous sections, we also investigate other adversary modeling strategies based on the TRADES \cite{zhang2019theoretically} (Eq. (\ref{eq:1})) baseline. Specifically, we provide the robustness results when conducting feature-level augmentation with random statistics or the target statistics without the uncertainty estimation, as shown in Table \ref{tab:7}. It can be seen that our distributional adversary modeling with the target statistics performs better than the random modeling. The full \modname{} configuration achieves the highest clean and robust accuracies among the modeling strategies in this table. Under AutoAttack, the full \modname{} configuration achieves $50.49\%$ accuracy, compared with $49.03\%$ for the intermediate-adversary variant and $49.19\%$ for the historical-adversary variant.

\begin{table}[!t]
	\centering
	\caption{Comparison of uncertainty estimation on diverse sample sets using ResNet-18 on CIFAR-10. We report both the accuracy (\%) and the average training time (min/epoch).}
	\vspace{-2mm}
	\renewcommand{\arraystretch}{0.5}
	\resizebox{0.9\linewidth}{!}{
		\begin{tabular}{ccccc}
			\toprule
			Sample Set & Clean & PGD & AA & Time (m) \\
			\midrule
			w/o Uncertainty & 82.45 & 52.21 & 48.88 & 2.58 \\
			Random Start (3 Samples) & 84.16 & 54.27 & 49.24 & 6.93 \\
			Random Start (5 Samples) & 84.80 & 54.49 & 49.79 & 11.30 \\
			Random Start (10 Samples) & 85.18 & 54.97 & 50.43 & 21.99 \\
			Random Start (15 Samples) & 85.07 & \textbf{55.05} & \textbf{50.57} & 32.78 \\
			\rowcolor{LightCyan}\textbf{EvoAUM} (Ours) & \textbf{85.25} & 54.88 & 50.41 & 3.20 \\
			
			\bottomrule
		\end{tabular}
		}
	\vspace{-3mm}
	\label{tab:8}
\end{table}

\noindent\textbf{Evolutionary search versus multi-restart sampling.}
To assess the effectiveness and efficiency of \modname{}, we compare evolutionary feature search with the random-start strategy for generating multiple pixel-space adversaries. As shown in Table \ref{tab:8}, we report both the performance and the average training time (min/epoch) of multi-restart sampling and \modname{}. We can easily observe a performance improvement on both clean and robust accuracy when increasing the size of the adversarial sample set via the random start strategy. However, this is also associated with a sharp increase of the time cost. \modname{} obtains comparable robustness by evaluating and selecting candidates in feature-statistic space, guided by the current PGD trajectory and warm-started by stored strategy parameters. It does not generate additional pixel-space PGD restarts or reuse historical adversarial samples.

\begin{table}[!t]
	\centering
	\caption{Impact (Acc.) of the position for adversarial uncertainty modeling using ResNet-18 on CIFAR-10.}
	\vspace{-2mm}
	\renewcommand{\arraystretch}{0.5}
	\resizebox{0.9\linewidth}{!}{
		\begin{tabular}{cccc}
			\toprule
			Position for Uncertainty Modeling & Clean & PGD & AA \\
			\midrule
			1st Convolutional Block & 86.04 & 54.92 & 50.83 \\
			\rowcolor{LightCyan}2nd Convolutional Block & 86.83 & 55.78 & 52.02 \\
			3rd Convolutional Block & 85.41 & 55.49 & 51.30 \\
			4th Convolutional Block & 86.17 & 55.28 & 51.21 \\
			Penultimate Layer & 86.29 & 55.12 & 51.64 \\
			
			\bottomrule
		\end{tabular}
	}
	\vspace{-1mm}
	\label{tab:supp_Pos_AUM}
\end{table}

\noindent\textbf{Different positions for \modname{}.} Since \modname{} is designed as a plug-and-play module that can be adapted to diverse levels of features, we here explore the efficacy of its positioning towards network robustness. As shown in Table \ref{tab:supp_Pos_AUM}, we report both clean and adversarially robust accuracy by using our proposed uncertainty feature augmentation across various layers of a ResNet-18 architecture, including the 1st to 4th convolutional blocks and the penultimate layer. We can observe that positioning the module in the middle layers, such as the 2nd convolutional block, or near the output, like the penultimate layer, significantly improves the model robustness against adversarial examples.

\noindent\textbf{Multi-step distributional alignment refinement.}
Table~\ref{tab:9} compares one, three, five, and ten steps of benign refinement (Eq.~\eqref{eq:8}) on CIFAR-10. One step gives $52.02\%$ AutoAttack accuracy at $3.97$ minutes per epoch, compared with $51.11\%$ and $5.86$ minutes for ten steps. These results support using one refinement step in our training configuration.

\begin{table}[!t]
	\centering
    \vspace{0mm}
	\caption{Comparison of benign refinement with diverse iterative steps using ResNet-18 on CIFAR-10. We report both the accuracy (\%) and the average training time (min/epoch).}
	\vspace{-2mm}
	\renewcommand{\arraystretch}{0.5}
	\resizebox{0.8\linewidth}{!}{
		\begin{tabular}{ccccc}
			\toprule
			Method & Clean & PGD & AA & Time (m) \\
			\midrule
			PGD-AT \cite{MadryMSTV18} & 83.80 & 51.40 & 47.68 & 2.95 \\
			TRADES \cite{zhang2019theoretically} & 82.45 & 52.21 & 48.88 & 3.28 \\
			MART \cite{0001ZY0MG20} & 82.20 & 53.94 & 48.04 & 3.78 \\
			HAT \cite{RadeM22} & 84.86 & 52.04 & 48.85 & 4.16 \\
			APT \cite{10177878} & 83.47 & 54.24 & 50.06 & 4.35 \\
			UIAT \cite{dong2023enemy} & 85.01 & 54.63 & 49.11 & 3.13 \\
			ARoW \cite{yang2023improving} & 83.46 & 54.07 & 50.29 & 3.56 \\
			\midrule
			\textbf{EvoUAD-AT} (3 Steps) & 86.34 & 55.74 & 51.87 & 4.58 \\
			\textbf{EvoUAD-AT} (5 Steps) & 86.36 & \textbf{55.85} & 51.80 & 4.92 \\
			\textbf{EvoUAD-AT} (10 Steps) & 86.18 & 55.31 & 51.11 & 5.86 \\
			\rowcolor{LightCyan}\textbf{EvoUAD-AT} (Single Step) & \textbf{86.83} & 55.78 & \textbf{52.02} & 3.97 \\
			
			\bottomrule
		\end{tabular}
	}
	\vspace{-2mm}
	\label{tab:9}
\end{table}

\subsection{Trade-off}
A well-established study has demonstrated the underlying trade-off between natural performance and adversarial robustness in the context of adversarial training \cite{zhang2019theoretically, RadeM22, dong2023enemy}. Here, we investigate the influence of the hyper-parameter $\beta$ in Eq. (\ref{eq:9}) for distribution-to-distribution prediction alignment, which plays a pivotal role in modulation the trade-off between the accuracies on clean samples and their adversarial counterparts. As depicted in Figure \ref{fig:4}, we can observe that an increase in $\beta$ correlates with the enhanced robust accuracy at the cost of the natural performance. Correspondingly, reducing $\beta$ yields an improvement in clean accuracy, conversely affecting robust accuracy. Our \methodname{} method retains the clean-classification and weighted-alignment structure of TRADES~\cite{zhang2019theoretically}, with elite adversarial features and a stochastic benign reference in the alignment term. Accordingly, $\beta$ balances clean classification and distributional prediction alignment in Eq.~\eqref{eq:9}. More analyses of hyper-parameters can be found in Section \ref{sec:hyperparas}.

\begin{figure}[!t]
	\centering
	\includegraphics[width=0.8\linewidth]{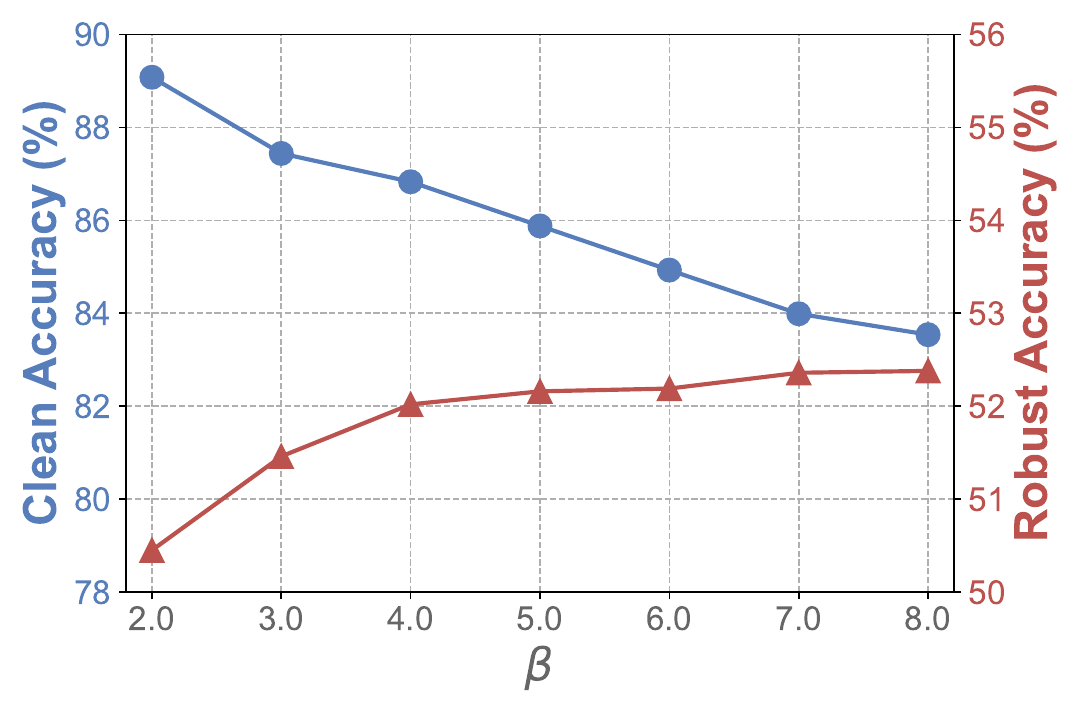}
	\vspace{-3mm}
	\caption{Accuracy-robustness trade-off of our EvoUAD-AT method by tuning the hyper-parameter $\beta$. We report both clean accuracy (\%) and (Auto-Attack) robust accuracy (\%).}
	\vspace{-1mm}
	\label{fig:4}
\end{figure}

\subsection{Visualization}

\noindent\textbf{Augmented features \textit{vs.} original features.}
Below, we explore the implicit correlation between our augmented adversarial features with the original adversarial features generated via the random start strategy. Specifically, we provide t-SNE visualizations of both the augmented and original features in Figure \ref{fig:5}, following the experimental setup in Figure \ref{fig:2}.

\begin{figure}[!t]
	\centering
	\begin{subfigure}[t]{0.32\linewidth} 
		\centering
		\includegraphics[width=1\linewidth]{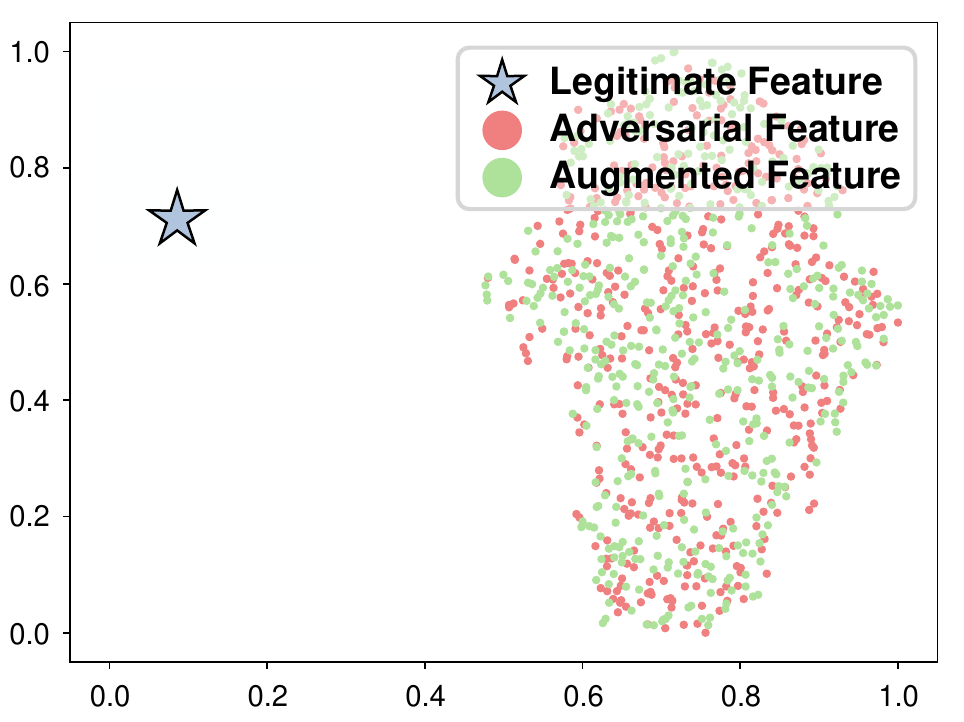}
		\vspace{-6mm}
		\caption{}
		\label{fig:5_1}
	\end{subfigure} 
	\begin{subfigure}[t]{0.32\linewidth}
		\centering
		\includegraphics[width=1\linewidth]{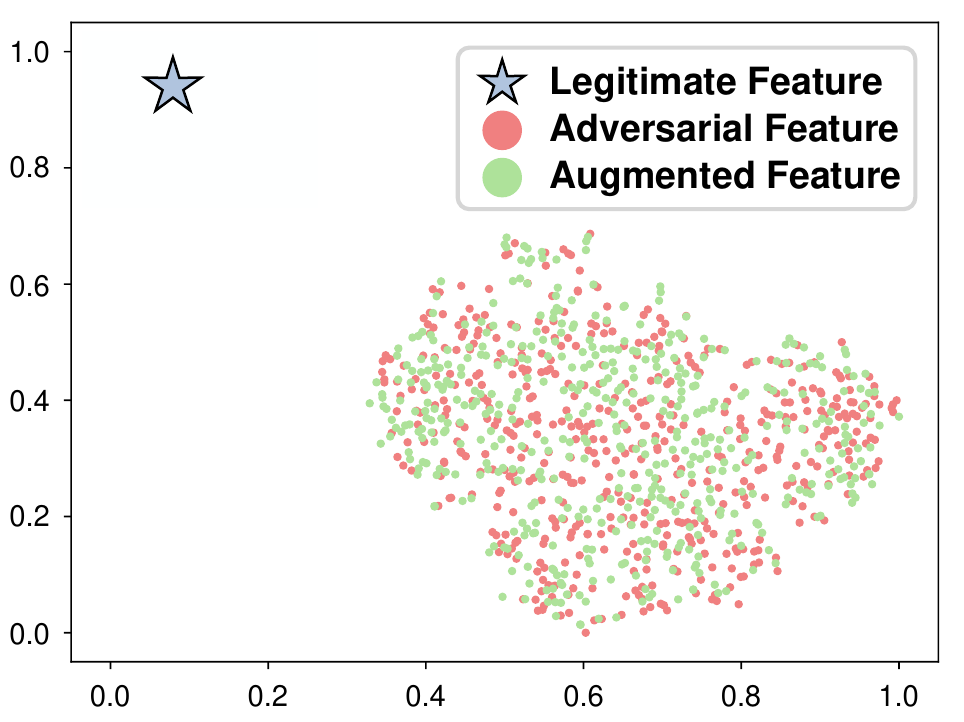}
		\vspace{-6mm}
		\caption{}
		\label{fig:5_2}
	\end{subfigure}
	\begin{subfigure}[t]{0.32\linewidth}
		\centering
		\includegraphics[width=1\linewidth]{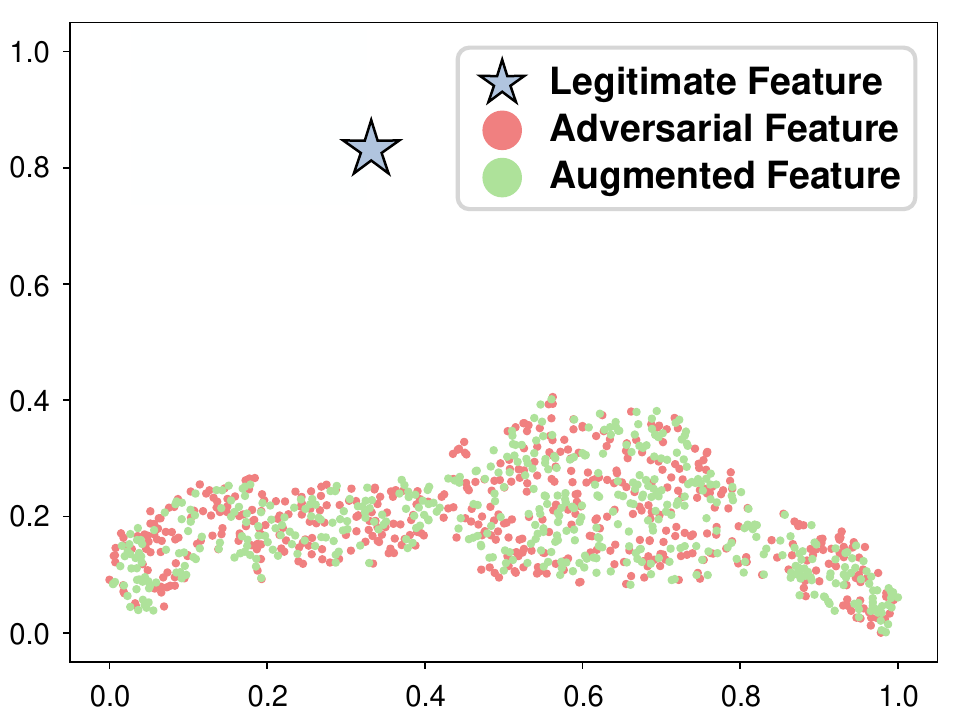}
		\vspace{-6mm}
		\caption{}
		\label{fig:5_3}
	\end{subfigure}
	\begin{subfigure}[t]{0.32\linewidth} 
		\centering
		\includegraphics[width=1\linewidth]{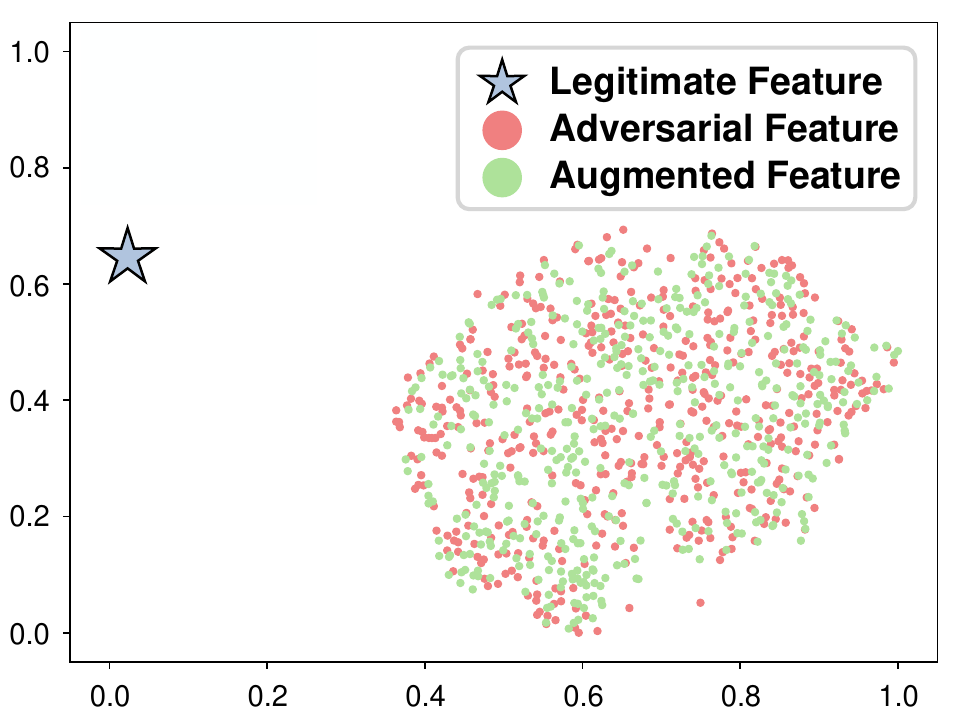}
		\vspace{-6mm}
		\caption{}
		\label{fig:5_4}
	\end{subfigure} 
	\begin{subfigure}[t]{0.32\linewidth}
		\centering
		\includegraphics[width=1\linewidth]{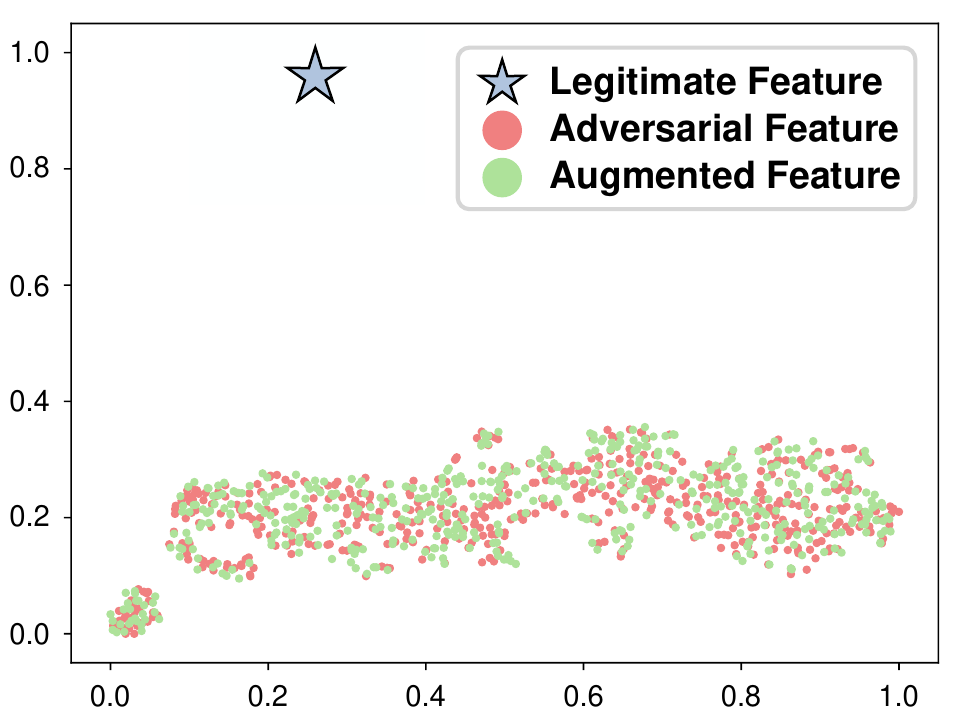}
		\vspace{-6mm}
		\caption{}
		\label{fig:5_5}
	\end{subfigure}
	\begin{subfigure}[t]{0.32\linewidth}
		\centering
		\includegraphics[width=1\linewidth]{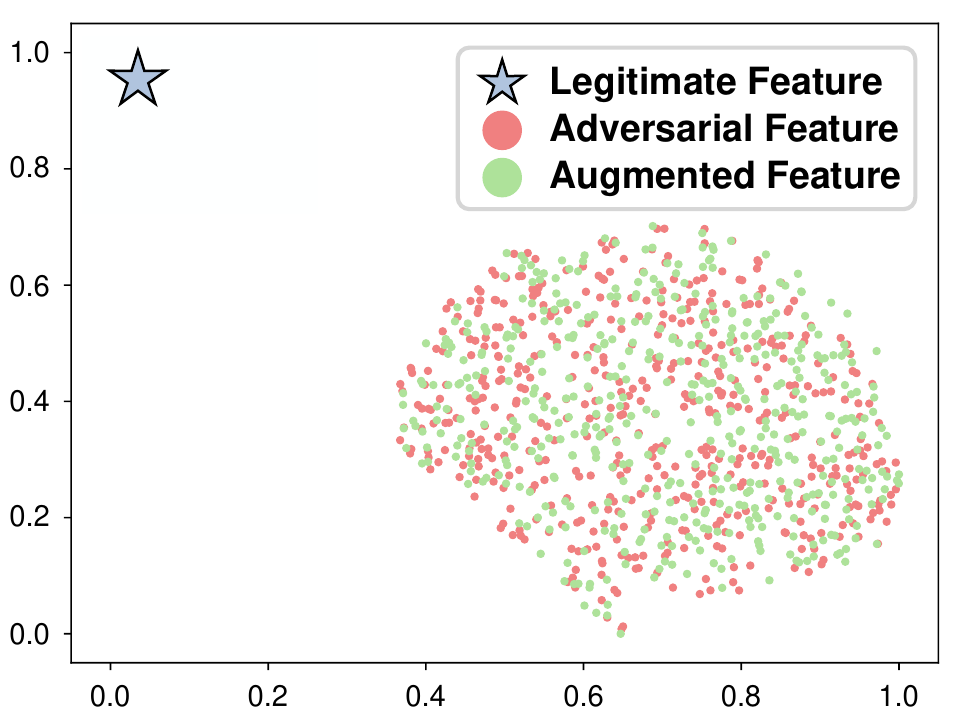}
		\vspace{-6mm}
		\caption{}
		\label{fig:5_6}
	\end{subfigure}
	\vspace{-2mm}
	\caption{t-SNE visualization of the legitimate feature, adversarial features, and augmented features on CIFAR-10 \cite{krizhevsky2009learning}. Each figure refers to a specific clean input.
	}
	\label{fig:5}
	\vspace{-1mm}
\end{figure}

It can be seen that a significant overlap occurs between the original adversarial feature distribution and our augmented counterpart. Such a feature-level overlapping also underscore the effectiveness of our adversarial uncertainty modeling approach in capturing the adversarial characteristics.

\noindent\textbf{Attention visualizations.}
As illustrated in Figure \ref{fig:6}, we present class-wise activation maps of our EvoUAD-AT method and other adversarial training approaches. Note that the attention maps are obtained by Grad-CAM \cite{selvaraju2017grad}, which relies on the gradient information flowing into the last convolutional layer of the target model. All the adversarial examples are generated via the PGD method with the perturbation radius $\epsilon=8/255$. It can be seen that our method achieves comprehensive attention regions that cover the entire target objects even when confronting unforeseen adversarial examples, which benefits from our uncertainty-aware distributional modeling.

\begin{figure}[!tb]
	\centering
	\includegraphics[width=0.95\linewidth]{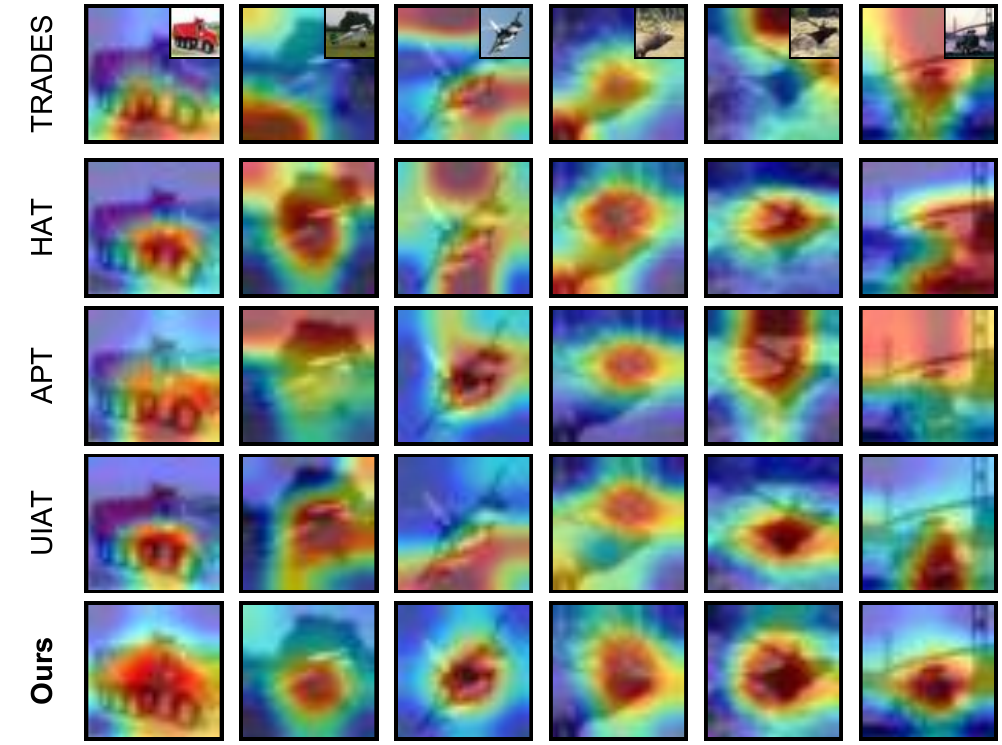}
	\vspace{-1mm}
	\caption{Grad-CAM visualizations of adversaries against diverse models on CIFAR-10 using ResNet-18.}
	\vspace{-1mm}
	\label{fig:6}
\end{figure}

\noindent\textbf{Robust overfitting.}
In addition to the accuracy-robustness trade-off, adversarial training has also been demonstrated to suffer from robust overfitting \cite{rice2020overfitting}, a significant issue that leads to an irreversible robustness drop on the test set following a certain number of training epochs. For a comprehensive understanding of our method, we here explore the learning curves of EvoUAD-AT and the standard adversarial training approach (PGD-AT) \cite{MadryMSTV18} in Figure \ref{fig:7}. For enhanced clarity and fairness, we extend the duration of training to $200$ epochs with the linear learning rate scheduling strategy.

\begin{figure}[!t]
	\centering
	\begin{subfigure}[t]{0.49\linewidth} 
		\centering
		\includegraphics[width=1\linewidth]{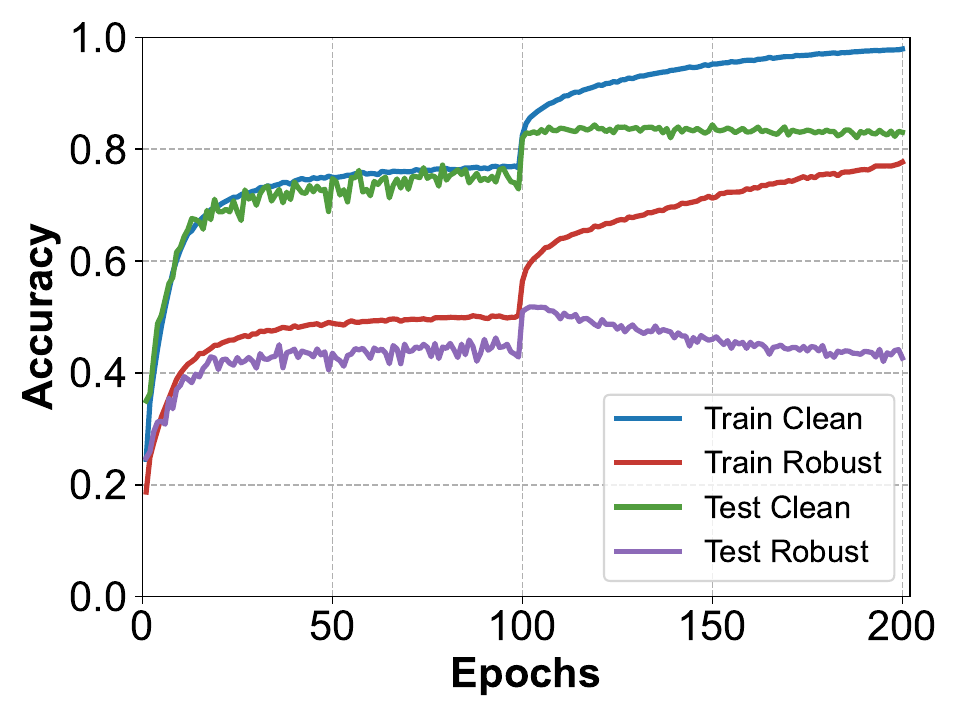}
		\vspace{-6mm}
		\caption{PGD-AT \cite{MadryMSTV18}}
		\label{fig:7_1}
	\end{subfigure} 
	\begin{subfigure}[t]{0.49\linewidth}
		\centering
		\includegraphics[width=1\linewidth]{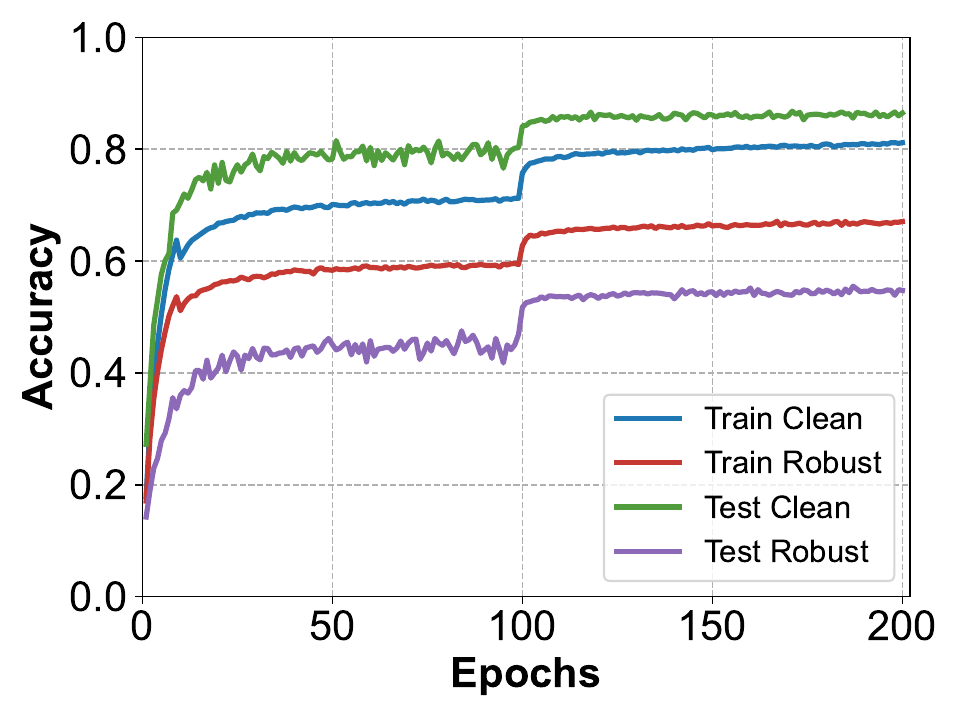}
		\vspace{-6mm}
		\caption{\textbf{Ours}}
		\label{fig:7_2}
	\end{subfigure}
	\vspace{-2mm}
	\caption{Learning curves of (a) PGD-AT \cite{MadryMSTV18} and (b) our method on CIFAR-10. We report both clean and (PGD-20) robust accuracies (\%) on the training and test sets.
	}
    \vspace{-1mm}
	\label{fig:7}
\end{figure}

It can be seen that PGD-AT highly suffers from robustness overfitting, while our method can effectively alleviate this issue. Our resilience against robust overfitting may be attributed to insights from \cite{rebuffi2021fixing}, which identified that such overfitting can be mitigated by data augmentation with external data for better generalizability. Our EvoUAD-AT effectively augments adversaries via an uncertainty-aware distributional modeling scheme with no need to access external data, thereby countering robust overfitting.

\subsection{Hyper-parameter Analysis}
\label{sec:hyperparas}
Here, we conduct hyper-parameter analysis for our method. Figure~\ref{fig:8} shows how clean and AutoAttack robust accuracy vary with the two loss weights in EvoUAD-AT. In the $\lambda_1$ sweep, AutoAttack accuracy peaks at $1.0$ and clean accuracy peaks at $1.5$. 
Increasing $\lambda_2$ from $0.05$ to $0.1$ raises AutoAttack accuracy while reducing clean accuracy. We use $\lambda_1=1.0$ and $\lambda_2=0.05$ in the main experiments.

\begin{figure}[!t]
	\centering
	\begin{subfigure}[t]{0.49\linewidth} 
		\centering
		\includegraphics[width=1\linewidth]{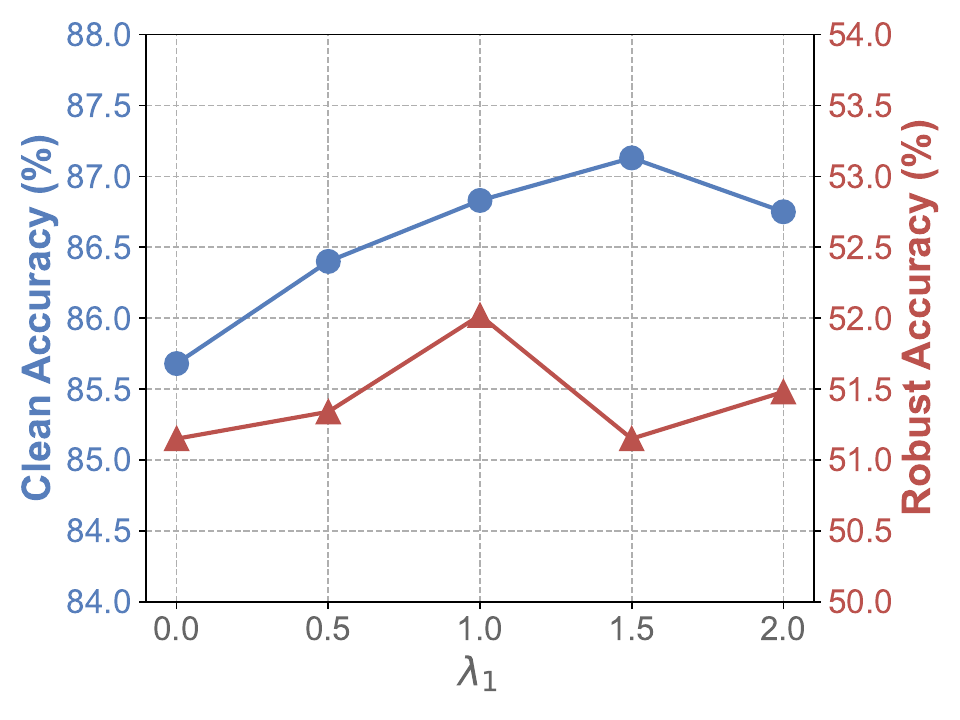}
		\vspace{-7mm}
		\caption{}
		\label{fig:8_1}
	\end{subfigure} 
	\begin{subfigure}[t]{0.49\linewidth}
		\centering
		\includegraphics[width=1\linewidth]{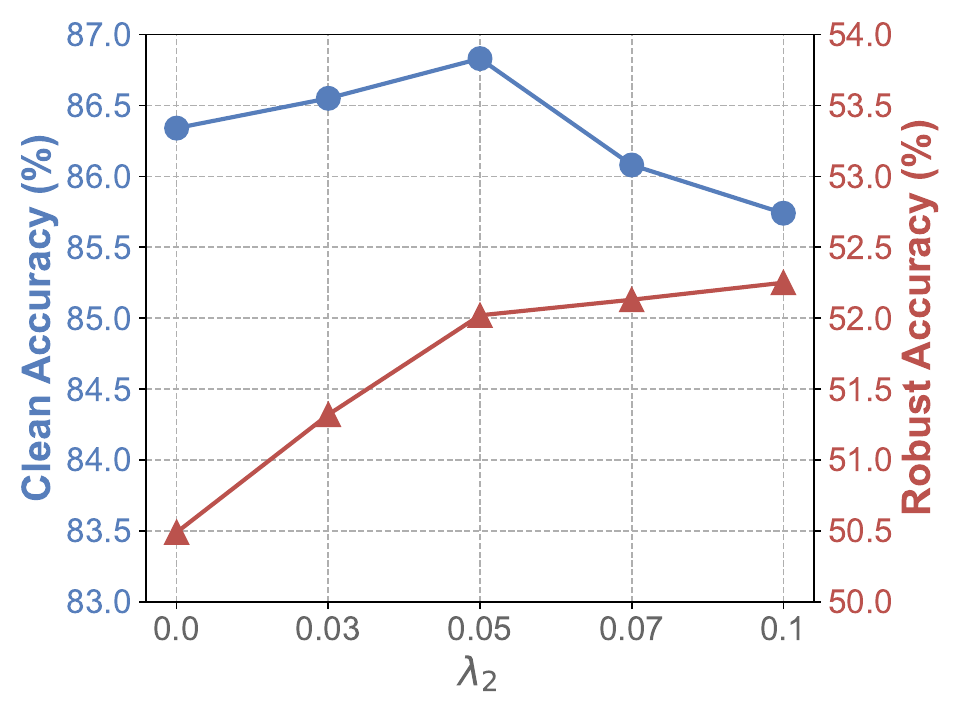}
		\vspace{-7mm}
		\caption{}
		\label{fig:8_2}
	\end{subfigure}
	\vspace{-3mm}
	\caption{Hyper-parameter sensitivity of EvoUAD-AT on clean and robust accuracies on CIFAR-10. We report the results of tuning factors $\lambda_1$ and $\lambda_2$ in (a) and (b), respectively.}
	\label{fig:8}
	\vspace{-1mm}
\end{figure}

\subsection{Why is Our \methodname{} Effective?}
We first examine how uncertainty-aware distributional augmentation affects the response of a network to adversarial perturbations. Following the setting in Figure~\ref{fig:1}, we compare the increases in feature variance and input-gradient norm from clean to adversarial inputs. Figure~\ref{fig:9} reports this comparison for EvoUAD-AT and TRADES~\cite{zhang2019theoretically}. At all four attack radii, both quantities increase less under EvoUAD-AT. For example, at $\epsilon=8/255$, feature variance increases by $14.3\%$ for EvoUAD-AT and $18.5\%$ for TRADES, while the corresponding increases in gradient norm are $75.3\%$ and $90.6\%$. This observation motivates modeling a range of adversarial features. In \methodname{}, \modname{} further uses the current PGD trajectory to guide the search and selects high-fitness features for training. Table~\ref{tab:1} reports the clean and robust accuracies of the resulting model.

\begin{figure}[!t]
	\centering
	\begin{subfigure}[t]{0.49\linewidth} 
		\centering
		\includegraphics[width=1\linewidth]{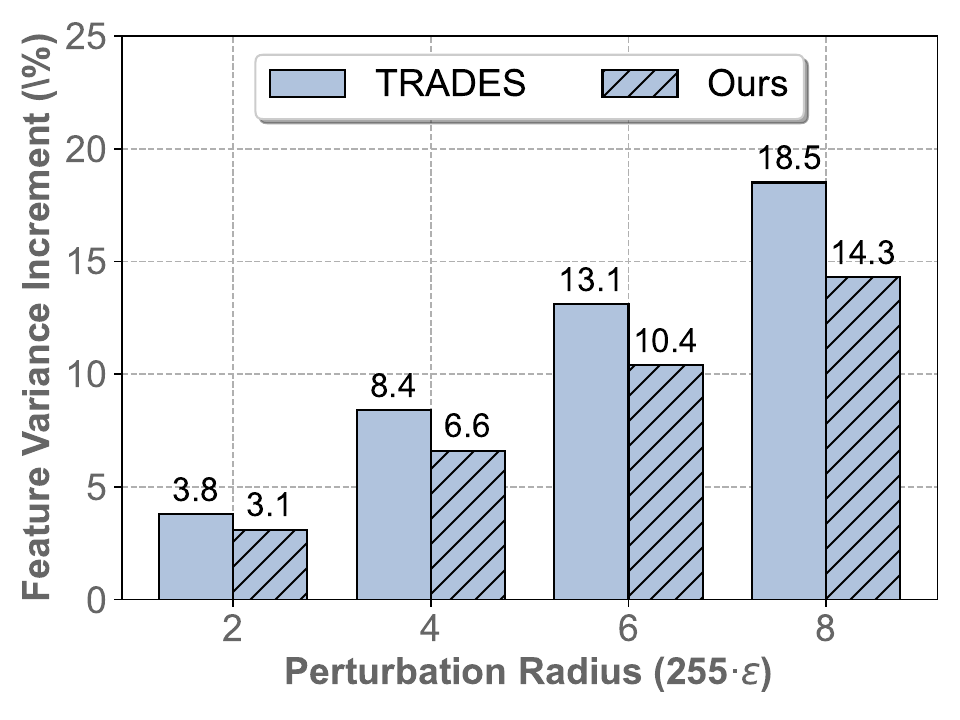}
		\vspace{-6mm}
		\caption{Feature Variance}
		\label{fig:9_1}
	\end{subfigure} 
	\begin{subfigure}[t]{0.49\linewidth}
		\centering
		\includegraphics[width=1\linewidth]{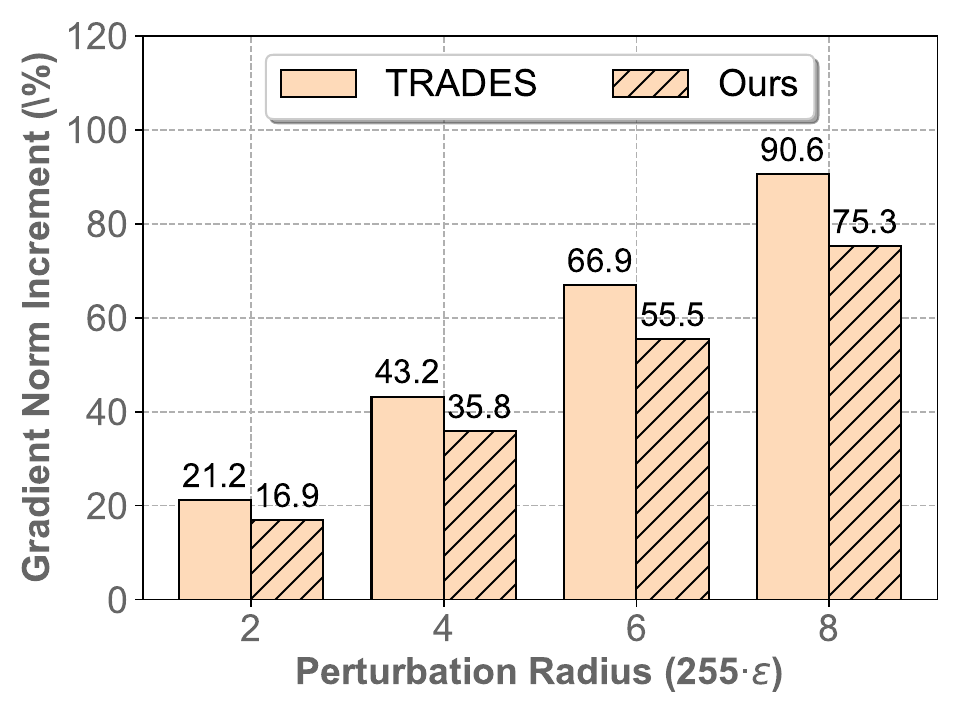}
		\vspace{-6mm}
		\caption{Gradient Norm}
		\label{fig:9_2}
	\end{subfigure}
	\vspace{-2mm}
	\caption{Increases from clean to adversarial inputs in (a) feature variance and (b) input-gradient norm across attack radii. The solid bars show TRADES, and the hatched bars show ours.}
	\label{fig:9}
	\vspace{-1mm}
\end{figure}

We next examine how the robustness difference varies with attack strength. Figure~\ref{fig:10} compares EvoUAD-AT and TRADES on CIFAR-10 when both models are trained at $\epsilon=8/255$. EvoUAD-AT achieves higher clean accuracy and higher AutoAttack accuracy at the tested positive radii up to $14/255$. At $16/255$, TRADES leads by $0.76$ percentage points. To investigate robustness when the stronger perturbation is also used during training, Table~\ref{tab:3} compares \methodname{} and TRADES at matched training and test radii. When both radii are $16/255$, \methodname{} achieves $22.59\%$ AutoAttack accuracy, compared with $18.70\%$ for TRADES.

\begin{figure}[!t]
	\centering
	\includegraphics[width=0.65\linewidth]{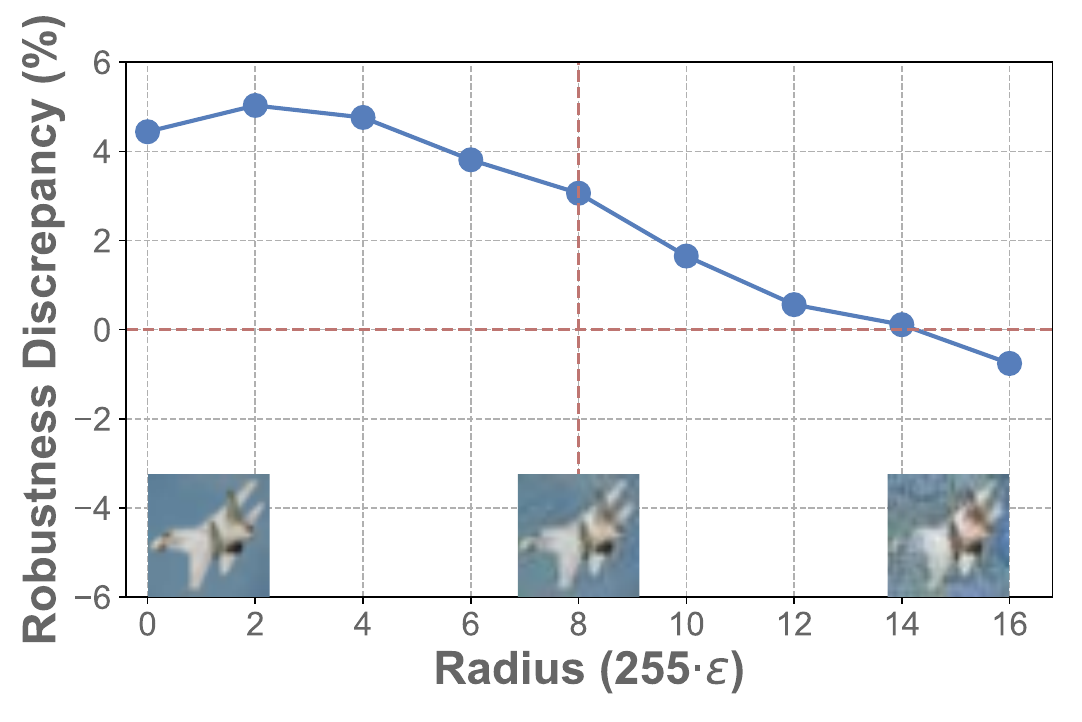}
	\vspace{-3mm}
	\caption{CIFAR-10 accuracy of EvoUAD-AT minus that of TRADES~\cite{zhang2019theoretically}, in percentage points. Both models were trained at $\epsilon=8/255$ (dashed line). Radius zero gives clean accuracy. Positive radii give AutoAttack accuracy. The insets show the clean image and adversarial examples at $8/255$ and $16/255$.}
	\vspace{-1mm}
	\label{fig:10}
\end{figure}

\section{Conclusion}
In this paper, we thoroughly investigate the limited generalization ability towards unforeseen adversarial samples stemming from point-by-point adversary generation and the side effects implicitly brought by adversaries. To mitigate such potential overfitting, we propose a novel distributional adversary modeling method that leverages the statistical information of adversaries with its uncertainty estimation for improved adversarial training. Considering the negative effect of misclassification on the prediction alignment reference, we design a distributional refinement scheme to correct each alignment reference, which reframes adversarial training within a distribution-to-distribution alignment paradigm. Furthermore, we develop an introspective gradient matching mechanism to enforce prediction invariance without accessing external models. Extensive experiments and in-depth analyses demonstrate the effectiveness and generalization ability of our proposed method across various settings and datasets.

\bibliographystyle{IEEEtran}
\bibliography{reference}

@String(CVPR= {IEEE Conf. Comput. Vis. Pattern Recog.})

@String(ICCV= {Int. Conf. Comput. Vis.})

@String(NIPS= {Adv. Neural Inform. Process. Syst.})

@String(ICLR = {Int. Conf. Learn. Represent.})

@String(CVPR  = {CVPR})

@String(ICCV  = {ICCV})

@String(NIPS  = {NeurIPS})

@String(ICLR  = {ICLR})

@inproceedings{GoodfellowSS15,
	author    = {Ian Goodfellow and
	Jonathon Shlens and
	Christian Szegedy},
	title     = {Explaining and Harnessing Adversarial Examples},
	booktitle = {ICLR},
	year      = {2015},
}

@inproceedings{SzegedyZSBEGF13,
	author = {Szegedy, Christian and Zaremba, Wojciech and Sutskever, Ilya and Bruna, Joan and Erhan, Dumitru and Goodfellow, Ian and Fergus, Rob},
	booktitle = {ICLR},
	title = {Intriguing properties of neural networks},
	year = {2014}
}

@inproceedings{MadryMSTV18,
  author       = {Aleksander Madry and
                  Aleksandar Makelov and
                  Ludwig Schmidt and
                  Dimitris Tsipras and
                  Adrian Vladu},
  title        = {Towards Deep Learning Models Resistant to Adversarial Attacks},
  booktitle    = {International Conference on Learning Representations},
  year         = {2018}
}

@inproceedings{zhang2019theoretically,
	title={Theoretically principled trade-off between robustness and accuracy},
	author={Zhang, Hongyang and Yu, Yaodong and Jiao, Jiantao and Xing, Eric and El Ghaoui, Laurent and Jordan, Michael},
	booktitle={ICML},
	pages={7472--7482},
	year={2019}
}

@inproceedings{carlini2017towards,
	title={Towards evaluating the robustness of neural networks},
	author={Carlini, Nicholas and Wagner, David},
	booktitle={2017 ieee symposium on security and privacy (sp)},
	pages={39--57},
	year={2017},
	organization={IEEE}
}

@inproceedings{croce2020reliable,
	title={Reliable evaluation of adversarial robustness with an ensemble of diverse parameter-free attacks},
	author={Croce, Francesco and Hein, Matthias},
	booktitle={International conference on machine learning},
	pages={2206--2216},
	year={2020},
	organization={PMLR}
}

@article{krizhevsky2009learning,
	title={Learning multiple layers of features from tiny images},
	author={Krizhevsky, Alex and Hinton, Geoffrey and others},
	year={2009},
	publisher={Citeseer}
}

@inproceedings{CroceASDFCM021,
  author    = {Francesco Croce and
               Maksym Andriushchenko and
               Vikash Sehwag and
               Edoardo Debenedetti and
               Nicolas Flammarion and
               Mung Chiang and
               Prateek Mittal and
               Matthias Hein},
  title     = {RobustBench: a standardized adversarial robustness benchmark},
  booktitle = {NeurIPS},
  year      = {2021}
}

@inproceedings{RadeM22,
  author       = {Rahul Rade and
                  Seyed{-}Mohsen Moosavi{-}Dezfooli},
  title        = {Reducing Excessive Margin to Achieve a Better Accuracy vs. Robustness Trade-off},
  booktitle    = {International Conference on Learning Representations},
  year         = {2022}
}

@inproceedings{xie2020adversarial,
  title={Adversarial examples improve image recognition},
  author={Xie, Cihang and Tan, Mingxing and Gong, Boqing and Wang, Jiang and Yuille, Alan L and Le, Quoc V},
  booktitle={Proceedings of the IEEE/CVF conference on CVPR},
  pages={819--828},
  year={2020}
}

@ARTICLE{10177878,
  author={Dong, Junhao and Yang, Lingxiao and Wang, Yuan and Xie, Xiaohua and Lai, Jianhuang},
  journal={IEEE Transactions on Image Processing}, 
  title={Toward Intrinsic Adversarial Robustness Through Probabilistic Training}, 
  year={2023},
  volume={32},
  number={},
  pages={3862-3872},
  doi={10.1109/TIP.2023.3290532}
}

@inproceedings{pang2022robustness,
  title={Robustness and accuracy could be reconcilable by (proper) definition},
  author={Pang, Tianyu and Lin, Min and Yang, Xiao and Zhu, Jun and Yan, Shuicheng},
  booktitle={International Conference on Machine Learning},
  year={2022}
}

@inproceedings{deng2009imagenet,
  title={Imagenet: A large-scale hierarchical image database},
  author={Deng, Jia and Dong, Wei and Socher, Richard and Li, Li-Jia and Li, Kai and Fei-Fei, Li},
  booktitle={2009 IEEE conference on computer vision and pattern recognition},
  pages={248--255},
  year={2009},
  organization={Ieee}
}

@article{ho2020denoising,
  title={Denoising diffusion probabilistic models},
  author={Ho, Jonathan and Jain, Ajay and Abbeel, Pieter},
  journal={Advances in NeurIPS},
  volume={33},
  pages={6840--6851},
  year={2020}
}

@inproceedings{WongRK20,
  author       = {Eric Wong and
                  Leslie Rice and
                  J. Zico Kolter},
  title        = {Fast is better than free: Revisiting adversarial training},
  booktitle    = {8th International Conference on Learning Representations, {ICLR}},
  year         = {2020}
}

@article{andriushchenko2020understanding,
  title={Understanding and improving fast adversarial training},
  author={Andriushchenko, Maksym and Flammarion, Nicolas},
  journal={Advances in Neural Information Processing Systems},
  volume={33},
  pages={16048--16059},
  year={2020}
}

@article{de2022make,
  title={Make some noise: Reliable and efficient single-step adversarial training},
  author={de Jorge Aranda, Pau and Bibi, Adel and Volpi, Riccardo and Sanyal, Amartya and Torr, Philip and Rogez, Gr{\'e}gory and Dokania, Puneet},
  journal={Advances in Neural Information Processing Systems},
  volume={35},
  pages={12881--12893},
  year={2022}
}

@inproceedings{0001ZY0MG20,
  author       = {Yisen Wang and
                  Difan Zou and
                  Jinfeng Yi and
                  James Bailey and
                  Xingjun Ma and
                  Quanquan Gu},
  title        = {Improving Adversarial Robustness Requires Revisiting Misclassified Examples},
  booktitle    = {ICLR},
  year         = {2020}
}

@inproceedings{dong2023enemy,
  title={The Enemy of My Enemy is My Friend: Exploring Inverse Adversaries for Improving Adversarial Training},
  author={Dong, Junhao and Moosavi-Dezfooli, Seyed-Mohsen and Lai, Jianhuang and Xie, Xiaohua},
  booktitle={Proceedings of the IEEE/CVF Conference on Computer Vision and Pattern Recognition},
  year={2023}
}

@inproceedings{BuiLT0P22,
  author       = {Anh Tuan Bui and
                  Trung Le and
                  Quan Hung Tran and
                  He Zhao and
                  Dinh Q. Phung},
  title        = {A Unified Wasserstein Distributional Robustness Framework for Adversarial Training},
  booktitle    = {International Conference on Learning Representations},
  year         = {2022}
}

@inproceedings{LiDGLSD22,
  author       = {Xiaotong Li and
                  Yongxing Dai and
                  Yixiao Ge and
                  Jun Liu and
                  Ying Shan and
                  Lingyu Duan},
  title        = {Uncertainty Modeling for Out-of-Distribution Generalization},
  booktitle    = {ICLR},
  year         = {2022}
}

@inproceedings{shi2019probabilistic,
  title={Probabilistic face embeddings},
  author={Shi, Yichun and Jain, Anil K},
  booktitle={IEEE/CVF International Conference on Computer Vision},
  pages={6902--6911},
  year={2019}
}

@article{dong2020adversarial,
  title={Adversarial distributional training for robust deep learning},
  author={Dong, Yinpeng and Deng, Zhijie and Pang, Tianyu and Zhu, Jun and Su, Hang},
  journal={Advances in Neural Information Processing Systems},
  volume={33},
  pages={8270--8283},
  year={2020}
}

@ARTICLE{10354457,
  author={Zhao, Chenglong and Mei, Shibin and Ni, Bingbing and Yuan, Shengchao and Yu, Zhenbo and Wang, Jun},
  journal={IEEE Transactions on Pattern Analysis and Machine Intelligence}, 
  title={Variational Adversarial Defense : A Bayes Perspective for Adversarial Training}, 
  year={2023}
}

@inproceedings{huang2017arbitrary,
  title={Arbitrary style transfer in real-time with adaptive instance normalization},
  author={Huang, Xun and Belongie, Serge},
  booktitle={Proceedings of the IEEE international conference on computer vision},
  pages={1501--1510},
  year={2017}
}

@inproceedings{moosavi2019robustness,
  title={Robustness via curvature regularization, and vice versa},
  author={Moosavi-Dezfooli, Seyed-Mohsen and Fawzi, Alhussein and Uesato, Jonathan and Frossard, Pascal},
  booktitle={IEEE/CVF Conference on Computer Vision and Pattern Recognition},
  year={2019}
}

@inproceedings{netzer2011reading,
  title={Reading digits in natural images with unsupervised feature learning},
  author={Netzer, Yuval and Wang, Tao and Coates, Adam and Bissacco, Alessandro and Wu, Baolin and Ng, Andrew Y and others},
  booktitle={NIPS workshop on deep learning and unsupervised feature learning},
  volume={2011},
  number={5},
  pages={7},
  year={2011},
  organization={Granada, Spain}
}

@article{shafahi2019adversarial,
  title={Adversarial training for free!},
  author={Shafahi, Ali and Najibi, Mahyar and Ghiasi, Mohammad Amin and Xu, Zheng and Dickerson, John and Studer, Christoph and Davis, Larry S and Taylor, Gavin and Goldstein, Tom},
  journal={Advances in neural information processing systems},
  volume={32},
  year={2019}
}

@inproceedings{selvaraju2017grad,
  title={Grad-cam: Visual explanations from deep networks via gradient-based localization},
  author={Selvaraju, Ramprasaath R and Cogswell, Michael and Das, Abhishek and Vedantam, Ramakrishna and Parikh, Devi and Batra, Dhruv},
  booktitle={Proceedings of the IEEE international conference on computer vision},
  pages={618--626},
  year={2017}
}

@inproceedings{rice2020overfitting,
  title={Overfitting in adversarially robust deep learning},
  author={Rice, Leslie and Wong, Eric and Kolter, Zico},
  booktitle={International conference on machine learning},
  pages={8093--8104},
  year={2020},
  organization={PMLR}
}

@inproceedings{yang2023improving,
  title={Improving adversarial robustness by putting more regularizations on less robust samples},
  author={Yang, Dongyoon and Kong, Insung and Kim, Yongdai},
  booktitle={International Conference on Machine Learning},
  pages={39331--39348},
  year={2023}
}

@inproceedings{wang2023better,
  title={Better diffusion models further improve adversarial training},
  author={Wang, Zekai and Pang, Tianyu and Du, Chao and Lin, Min and Liu, Weiwei and Yan, Shuicheng},
  booktitle={International Conference on Machine Learning},
  pages={36246--36263},
  year={2023}
}

@inproceedings{HendrycksD19,
  author       = {Dan Hendrycks and
                  Thomas G. Dietterich},
  title        = {Benchmarking Neural Network Robustness to Common Corruptions and Perturbations},
  booktitle    = {ICLR},
  year         = {2019}
}

@article{rebuffi2021fixing,
  title={Fixing data augmentation to improve adversarial robustness},
  author={Rebuffi, Sylvestre-Alvise and Gowal, Sven and Calian, Dan A and Stimberg, Florian and Wiles, Olivia and Mann, Timothy},
  journal={arXiv preprint arXiv:2103.01946},
  year={2021}
}

@article{zhu2022toward,
  title={Toward understanding and boosting adversarial transferability from a distribution perspective},
  author={Zhu, Yao and Chen, Yuefeng and Li, Xiaodan and Chen, Kejiang and He, Yuan and Tian, Xiang and Zheng, Bolun and Chen, Yaowu and Huang, Qingming},
  journal={IEEE Transactions on Image Processing},
  volume={31},
  pages={6487--6501},
  year={2022},
  publisher={IEEE}
}

@article{su2024multi,
  title={Multi-Label Adversarial Attack With New Measures and Self-Paced Constraint Weighting},
  author={Su, Fengguang and Wu, Ou and Zhu, Weiyao},
  journal={IEEE Transactions on Image Processing},
  volume={33},
  pages={3809--3822},
  year={2024},
  publisher={IEEE}
}

@article{li2025map,
  title={Map: Masked adversarial perturbation for boosting black-box attack transferability},
  author={Li, Kaige and Wan, Maoxian and Geng, Qichuan and Shi, Weimin and Cao, Xiaochun and Zhou, Zhong},
  journal={IEEE Transactions on Image Processing},
  year={2025},
  publisher={IEEE}
}

@article{huang2023fast,
  title={Fast adversarial training with adaptive step size},
  author={Huang, Zhichao and Fan, Yanbo and Liu, Chen and Zhang, Weizhong and Zhang, Yong and Salzmann, Mathieu and S{\"u}sstrunk, Sabine and Wang, Jue},
  journal={IEEE Transactions on Image Processing},
  volume={32},
  pages={6102--6114},
  year={2023},
  publisher={IEEE}
}

@InProceedings{Wang_2026_CVPR,
    author    = {Wang, Yanyun and Ye, Qingqing and Liu, Li and Liang, Zi and Hu, Haibo},
    title     = {Robust Alignment: Harmonizing Clean Accuracy and Adversarial Robustness in Adversarial Training},
    booktitle = {Proceedings of the IEEE/CVF Conference on Computer Vision and Pattern Recognition (CVPR) Findings},
    month     = {June},
    year      = {2026},
    pages     = {768-778}
}

@InProceedings{Zhang_2025_ICCV,
    author    = {Zhang, Kejia and Weng, Juanjuan and Li, Shaozi and Luo, Zhiming},
    title     = {Towards Adversarial Robustness via Debiased High-Confidence Logit Alignment},
    booktitle = {Proceedings of the IEEE/CVF International Conference on Computer Vision (ICCV)},
    month     = {October},
    year      = {2025},
    pages     = {2783-2792}
}

@book{cover2006elements,
  title     = {Elements of Information Theory},
  author    = {Cover, Thomas M. and Thomas, Joy A.},
  edition   = {2nd},
  publisher = {Wiley},
  year      = {2006}
}

@misc{hansen2016tutorial,
  title         = {The {CMA} Evolution Strategy: A Tutorial},
  author        = {Hansen, Nikolaus},
  year          = {2016},
  eprint        = {1604.00772},
  archivePrefix = {arXiv},
  howpublished  = {arXiv:1604.00772}
}

@inproceedings{maheswaranathan2019guided,
  title     = {Guided Evolutionary Strategies: Augmenting Random Search with Surrogate Gradients},
  author    = {Maheswaranathan, Niru and Metz, Luke and Tucker, George and Choi, Dami and Sohl-Dickstein, Jascha},
  booktitle = {Proc. Int. Conf. Mach. Learn. (ICML)},
  pages     = {4264--4273},
  year      = {2019}
}

@article{massart1990dkw,
  title   = {The Tight Constant in the {D}voretzky--{K}iefer--{W}olfowitz Inequality},
  author  = {Massart, Pascal},
  journal = {Ann. Probab.},
  volume  = {18},
  number  = {3},
  pages   = {1269--1283},
  year    = {1990}
}

@article{rockafellar2002cvar,
  title   = {Conditional Value-at-Risk for General Loss Distributions},
  author  = {Rockafellar, R. Tyrrell and Uryasev, Stanislav},
  journal = {J. Banking Finance},
  volume  = {26},
  number  = {7},
  pages   = {1443--1471},
  year    = {2002}
}

@inproceedings{ros2008sep,
  title     = {A Simple Modification in {CMA-ES} Achieving Linear Time and Space Complexity},
  author    = {Ros, Raymond and Hansen, Nikolaus},
  booktitle = {Proc. Parallel Problem Solving from Nature (PPSN X)},
  pages     = {296--305},
  year      = {2008}
}

\end{document}